\documentclass[single-column]{fairmeta}
\usepackage[most]{tcolorbox}

\usepackage{amsmath}
\usepackage{amssymb}
\usepackage{xspace}

\usepackage{graphicx}
\usepackage{booktabs}
\usepackage{mfirstuc}
\usepackage{calc}
\usepackage{multirow}
\usepackage{array}

\usepackage{booktabs}
\usepackage{multirow}
\usepackage{pgfplotstable}
\pgfplotsset{compat=1.18}
\usepackage[table]{xcolor}
\definecolor{bestblue}{RGB}{226,240,255}
\definecolor{bestpink}{RGB}{255,228,236}

\usepackage{hyperref}

\ifdefined\showcomments

\else

\fi

\newcommand{\modelname}{Chameleon\xspace}
\newcommand{\benchname}{Camo-Dataset\xspace}

\newcommand{\serveplate}{Clean a specified plate\xspace}
\newcommand{\shellgame}{Play shell game\xspace}
\newcommand{\addpeasoning}{Add various seasonings\xspace}
\newcommand{\DP}{Diffusion Policy\xspace}

\newcommand{\modelparams}{$\sim$60M\xspace}

\title{Chameleon: Control-Indexed Prospective Memory for Visuomotor Manipulation}

\author[1, 2, *]{Xinying Guo}
\author[1, *]{Chenxi Jiang}
\author[1]{Hyun Bin Kim}
\author[3]{Yuhang Han}
\author[2]{Ying Sun}
\author[1]{Yang Xiao}
\author[1,\dagger]{Jianfei Yang}

\affiliation[1]{MARS Lab, Nanyang Technological University}
\affiliation[2]{Institute for Infocomm Research, A*STAR, Singapore}
\affiliation[3]{National University of Singapore}

\contribution[*]{Equal Contribution}
\contribution[\dagger]{Corresponding Author}

\abstract{
Robots often observe information that determines a future action long before that action is executed.
In a shell game, for example, a robot first sees which cup hides the ball, watches the cups move, and only later needs to choose the correct cup.
The final observation alone is not enough for a decision: the correct action depends on an earlier event.
We refer to this temporal gap as \emph{observation--action delay}.
It makes memory a policy-facing problem: a policy must keep similar histories distinct, retrieve the past event relevant to the current decision, and convert that recall into an action-ready state.
We call these requirements separability, addressability, and prospectiveness.
We introduce \textbf{Chameleon}, a \modelparams{} visuomotor policy for \emph{control-indexed prospective memory}.
Chameleon writes embodied event memory, preserves separable histories, retrieves control-relevant traces, and trains the resulting working state to be prospective.
We also introduce \textbf{Camo-Dataset}, a real-robot benchmark that isolates observation--action delay by making the decision scene visually ambiguous, so the correct action must be inferred from earlier observations.
Chameleon improves decision/end-to-end success on Camo-Dataset from 22.5\%/21.3\% to 80.8\%/71.3\%.
On public long-horizon memory benchmarks, it achieves 
87.1\%$\pm$0.8\% on LIBERO-10, 97.3\%$\pm$4.5\% on MemoryBench, and 75.1\%$\pm$1.4\% on MIKASA-Robo, setting the state of the art for same-size models and exceeding multiple larger VLA baselines under the reported protocols.
Probes and ablations show that Chameleon learns separable, addressable, and prospective memory, and that these properties drive its performance gains.
}

\correspondence{Jianfei Yang at \email{jianfei.yang@ntu.edu.sg}}
\metadata[Code]{\url{https://github.com/gxyes/MARS_Chameleon}}

\begin{document}

\maketitle

\section{Introduction}
\label{sec:intro}


\begin{figure}[t]
\centering
\includegraphics[page=1,width=\textwidth,trim=51bp 0bp 0bp 0bp,clip]{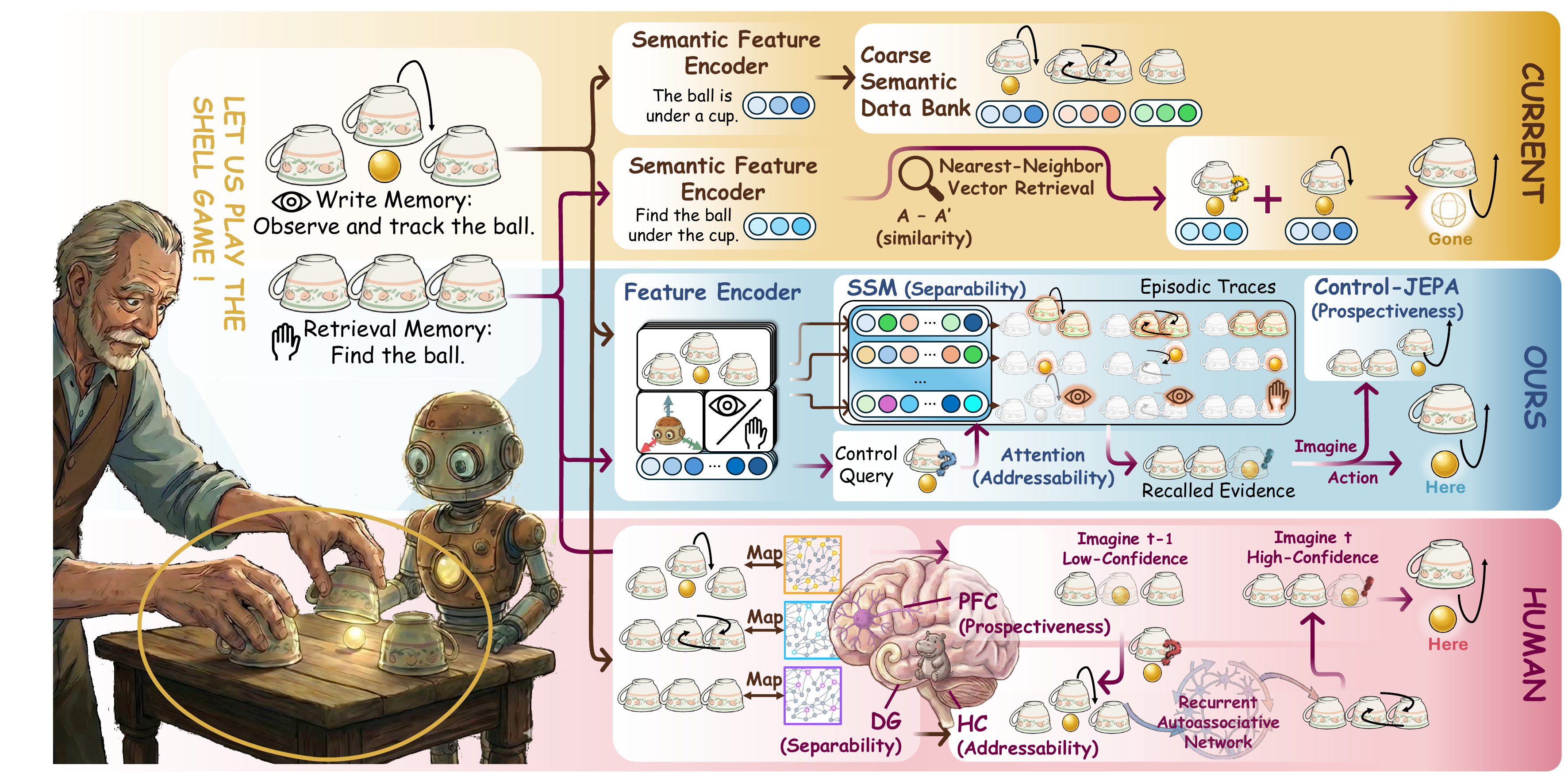}
\caption{\textbf{Observation--action delay requires control-indexed memory.}
\textbf{Current:} Semantic summaries or similarity-based visual retrieval can select a plausible but wrong trace.
\textbf{Ours \& Human:} Effective memory requires three capabilities: separability, addressability, and prospectiveness.}
\label{fig:motivation}
\vspace{-1em}
\end{figure}

Long-horizon and memory-dependent robot benchmarks reveal a recurring challenge: information observed earlier may become necessary for later action only after it is no longer directly observable~\cite{liu2023libero,cherepanov2025memory,fang2025sam2act}.
In a shell game, a robot sees which cup hides the ball, watches the cups shuffle, and later must choose the correct cup.
By the time it has to choose, the ball is no longer visible.
The correct action depends on what it saw earlier, not on the current observation alone.
We call this temporal gap between observing information and using it for action an \emph{observation--action delay}.
Such delays make manipulation non-Markovian in the policy's observation space: the same current input can require different actions depending on the history that produced it.

Observation--action delay is not solved by merely giving a policy more frames.
The policy needs memory that keeps action-relevant events available until the decision point and represents them in a form usable for control.
In the shell game, this means preserving which cup hid the ball and how that cup moved, not merely remembering that a ball was hidden.
Existing memory-augmented policies often use compact summaries or retrieve similar past observations.
Summaries can erase the fine-grained event trace needed for action, while similarity-based retrieval can return a visually plausible but control-irrelevant frame~\cite{lewis2020retrieval,zhu2024retrieval,anwar2025remembr}.
Even retrieving the right past observation is not enough: the policy must convert it into a state that can guide the current action.
Thus, embodied memory should not be defined only by what it stores or retrieves, but by whether it makes the past usable for control.

Human episodic memory offers a useful analogy for such policy-facing memory~\cite{allen2013evolution}, which we use as functional inspiration rather than a biological blueprint.
The dentate gyrus (DG) maps overlapping inputs to less overlapping codes, helping similar experiences be stored as distinct episodes~\cite{bakker2008pattern}.
This motivates \emph{separability}: perceptually similar histories should not collapse into the same memory state.
The interaction between the prefrontal cortex (PFC) and hippocampus (HC) supports cue-dependent retrieval of episodic memories~\cite{worsfold2025revisiting}.
This motivates \emph{addressability}: the policy should retrieve the trace relevant to the current decision.
PFC supports goal-directed prediction of remembered episodes, allowing recalled information to be interpreted in light of upcoming actions~\cite{zheng2025flexible}. 
This motivates \emph{prospectiveness}: the ability to consolidate recall into a working state that supports future action.
Together, these principles define policy-facing memory for delayed control: preserve the distinctions that matter, retrieve the trace the current decision asks for, and make that trace usable for future action.

We instantiate these principles in \modelname, a \modelparams-parameter visuomotor policy for \emph{control-indexed prospective memory}.
At each timestep, \modelname takes RGB views, proprioception, and an optional language instruction, and predicts a short horizon of future actions.
It writes the current moment into embodied event tokens that bind visual, proprioceptive, and language evidence.
These tokens are propagated through a slow episode-level memory module, preserving distinct histories over time rather than compressing them into a single state. This realizes separability.
When an action decision is needed, a learned control index derived from the current embodied state queries memory and retrieves the trace relevant to the present choice. This realizes addressability.
The retrieved trace is then consolidated into a fast working state for action prediction.
We train this state with Control-JEPA~\cite{assran2023selfsupervised} to predict future control context, making memory prospective rather than merely descriptive and providing a direct learning signal under observation--action delay.

Our evaluation follows a diagnostic-to-generalization-to-mechanism structure.
We first build \benchname, a real-robot UR5 benchmark that isolates observation--action delay and separates memory-dependent decision success from execution success, where \modelname improves decision/end-to-end success from 22.5\%/21.3\% to 80.8\%/71.3\%.
We then test whether the same design generalizes to public long-horizon memory benchmarks, including LIBERO-10~\cite{liu2023libero}, MemoryBench~\cite{fang2025sam2act}, and MIKASA-Robo~\cite{cherepanov2025memory}, where \modelname reaches state-of-the-art performance among same-size models and outperforms several larger VLA baselines.
Finally, representation probes and ablations show that separability, addressability, and prospectiveness are realized and functionally necessary.

Our contributions are threefold.
First, we formulate \emph{observation--action delay} as a policy-facing memory bottleneck in robot manipulation and identify separability, addressability, and prospectiveness as its key requirements.
Second, we introduce \modelname, a visuomotor policy for \emph{control-indexed prospective memory} that keeps event histories distinct, retrieves control-relevant traces, and converts recall into action-ready state.
Third, we introduce \benchname, a diagnostic real-robot benchmark for observation--action delay, and show that \modelname achieves state-of-the-art same-size performance across \benchname, public benchmarks, probes, and ablations.

\section{Related Work}


\paragraph{Embodied memory: methods and benchmarks.}
Embodied policies usually use a recurrent hidden state to keep experience~\cite{gu2022efficiently,gu2023mamba,dao2024mamba2}, extend the observation window~\cite{parisotto2020stabilizing,hawthorne2022general}, or store experience in an external memory bank~\cite{lewis2020retrieval,karpukhin2020dense,borgeaud2022improving,yao2022react,park2023generative}. 
Recent robot-memory systems have explored many forms of memory, including video-text memories, keyframe retrieval, scene or episode memories, spatial maps, semantic hierarchies, recovery modules, prompt memories, and object-centric state tracking~\cite{torne2026vlas,sridhar2025memer,lin2025echovla,qian2026escape,hu2026echo,zeng2026helm,shi2026memoryvla,li2025mapvla,wang2025karma,liu2024meia,xie2024embodied,zhu2024retrieval,mon2025embodied,anwar2025remembr,chung2025rethinking}. 
These systems establish that memory helps at different scales. 
\modelname addresses the question these mechanisms leave open: can a policy retrieve the trace that matters for the current control decision, rather than the most recent or most similar trace?
We use three public benchmarks: MemoryBench~\cite{fang2025sam2act} for spatial memory in manipulation, MIKASA-Robo~\cite{cherepanov2025memory} for simulated non-Markovian tasks, and LIBERO-10~\cite{liu2023libero} for language-conditioned long-horizon imitation. \benchname instead creates controlled real-robot perceptual aliasing, allowing memory mistakes to be separated from manipulation failures.


\section{Method}
\label{sec:method}

\begin{figure}[!htbp]
    \centering
    \includegraphics[page=1,width=\textwidth,trim=325bp 455bp 315bp 295bp,clip]{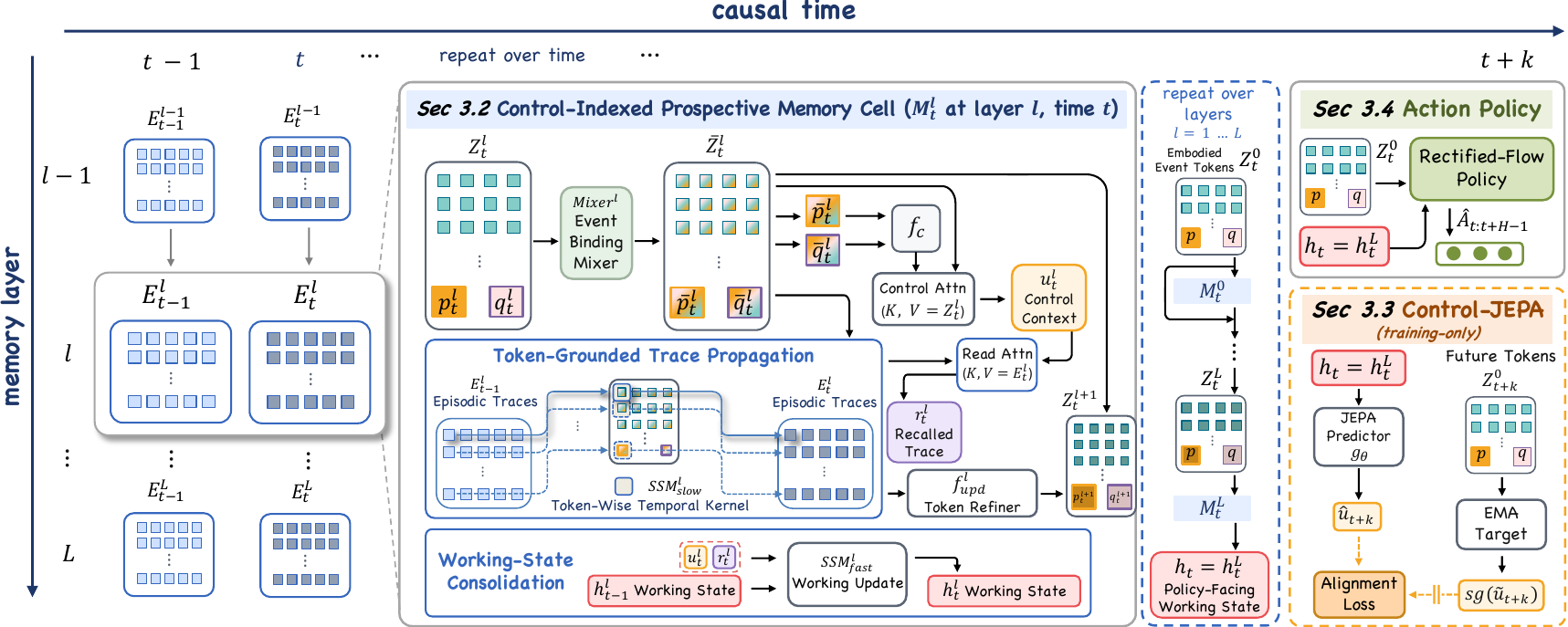}
    \caption{
    \textbf{Chameleon overview.} Chameleon writes embodied event tokens, propagates them as token-grounded traces, addresses them with a learned control index, and consolidates the recalled trace into the prospective policy state $h_t$. The state conditions a rectified-flow action policy, while Control-JEPA trains it to predict future control context.
    }
    \label{fig:method}
\end{figure}


\modelname{} is a causal visuomotor policy for \emph{control-indexed prospective memory}.
At time $t$, it receives only observations available up to the present and predicts a future action horizon:
\begin{equation}
    Z_t^0=\Phi_{\mathrm{evt}}(O_t,s_t,g_t),\qquad
    h_t=\Phi_{\mathrm{mem}}(Z_{\leq t}^{0}),\qquad
    \hat{A}_{t:t+H-1}=\Phi_{\mathrm{act}}(h_t).
    \label{eq:chameleon_pipeline}
\end{equation}
Here $O_t=\{I_t^v\}_{v=1}^{V}$ are RGB views, $s_t$ is proprioception, $g_t$ is an optional language instruction, $Z_t^0$ are input embodied event tokens, $h_t$ is the prospective working state, and $\hat{A}_{t:t+H-1}$ is the predicted action horizon.
Future observations are never used in inference. 
They appear only during training as targets for the prospective objective in Sec.~\ref{sec:method_control_jepa}.

The memory module follows the three requirements introduced in Sec.~\ref{sec:intro}.
To support \emph{separability}, \modelname{} writes each timestep as multiple localized embodied event tokens and propagates token-grounded traces rather than compressing history into a single recurrent vector.
To support \emph{addressability}, it forms a learned control index from the current embodied state and uses it to recall the trace relevant to the present decision.
To support \emph{prospectiveness}, it consolidates the recalled trace into a working state trained to predict future control context.
The resulting computation is a write--propagate--address--consolidate loop: write event evidence, propagate causal traces, address memory by the current control question, and consolidate recall into an action-ready state.

In implementation, visual observations are encoded with a DP-style patch encoder~\cite{chi2023diffusionpolicy}, language is encoded with a frozen DistilBERT text encoder~\cite{sanh2019distilbert}, token-grounded traces are propagated with selective state-space layers~\cite{gu2023mamba,dao2024mamba2}, and actions are generated by a transformer rectified-flow head~\cite{vaswani2017attention,lipman2022flow,liu2023flowstraight}.
Dimensions, token resolutions, horizon sets, and loss weights are reported in the Appendix.

\subsection{Embodied Event Tokens}
\label{sec:method_event_tokens}

Observation--action delay makes it difficult to know in advance which parts of an observation must be preserved for future control.
Chameleon therefore writes each timestep as a set of embodied event tokens, instead of immediately pooling the observation into a global descriptor.
For each camera view $v$, the visual encoder produces patch-level tokens:
\begin{equation}
    X_t^v = E_{\mathrm{vis}}^v(I_t^v)\in\mathbb{R}^{N_v\times d}.
    \label{eq:visual_tokens}
\end{equation}
Proprioception and language are projected into the same token space:
\begin{equation}
    p_t=E_{\mathrm{prop}}(s_t)\in\mathbb{R}^{d},\qquad
    q_t=E_{\mathrm{lang}}(g_t)\in\mathbb{R}^{d}.
    \label{eq:prop_lang_tokens}
\end{equation}
The language encoder is frozen, the projection layers are learned, and tasks without language use a learned null instruction token.
The embodied-event-token set is
\begin{equation}
    Z_t^{0}=\mathrm{Concat}\left[X_t^1,\ldots,X_t^V,p_t,q_t\right]\in\mathbb{R}^{N\times d}.
    \label{eq:event_tokens}
\end{equation}
Visual tokens encode local evidence, proprioception anchors the body, and language specifies the task. Together they form embodied event tokens for memory writing, propagation, addressing, and recall.

\subsection{Control-Indexed Prospective Memory}
\label{sec:method_memory}

The memory module applies $L$ causal layers to the sequence of embodied event tokens.
Each layer performs four operations: event binding, token-grounded trace propagation, control-indexed recall, and working-state consolidation.

\paragraph{Event binding.}
Each timestep, event tokens first interact through a residual self-attention block:
\begin{equation}
    \bar{Z}_t^\ell=\mathrm{Mixer}^{\ell}(Z_t^\ell).
    \label{eq:event_mixer}
\end{equation}
This is the write interface.
Before temporal propagation, visual, proprioceptive, and language tokens reinterpret one another, so the written event already contains task- and body-conditioned evidence.

\vspace{-0.5em}

\paragraph{Token-grounded trace propagation.} We use \emph{slow} to denote the episode-level memory timescale, in contrast to the \emph{fast} working state used for immediate action prediction.
A single recurrent state can merge similar histories, whereas delayed control requires preserving the distinctions that determine future actions.
Chameleon therefore propagates a bank of token-grounded traces $E^\ell$ across the causal history.
For each token stream $i$, a shared token-wise temporal kernel processes the causal sequence:
\begin{equation}
    e_{1:T,i}^\ell=
    \mathrm{SSM}_{\mathrm{slow}}^\ell
    \left(\bar{Z}_{1:T,i}^\ell\right),
    \qquad
    E_t^\ell=\{e_{t,i}^\ell\}_{i=1}^{N}.
    \label{eq:slow_episodic}
\end{equation}

\paragraph{Control index and control context.}
At decision time, the policy does not need all remembered evidence equally; it needs the trace that answers the current control question.
Each layer, therefore, forms a learned \emph{control index $c_t^\ell$} from the current embodied tokens. In the default setting, the control index is derived from the proprioceptive and language tokens after event binding:
\begin{equation}
    c_t^\ell = f_c^\ell\!\left([\bar{p}_t^\ell,\bar{q}_t^\ell]\right)\in\mathbb{R}^{d}.
    \label{eq:control_seed}
\end{equation}
Before reading memory, the current bound event tokens set refines the control index into a \emph{control context} $u_t^\ell$, which combines task, body, and present-scene evidence before addressing the trace bank.
\begin{equation}
    u_t^\ell=\mathrm{LN}\!\left(c_t^\ell+
    \mathrm{Attn}(c_t^\ell,\bar{Z}_t^\ell,\bar{Z}_t^\ell)\right).
    \label{eq:control_context}
\end{equation}
\paragraph{Control-indexed recall and working-state consolidation.}
The control context recalls remembered evidence by attending over token-grounded traces:
\begin{equation}
    r_t^\ell=
    \mathrm{LN}\!\left(
    \mathrm{Attn}(u_t^\ell,E_t^\ell,E_t^\ell)
    \right).
    \label{eq:episodic_recall}
\end{equation}
The recalled trace $r_t^\ell$ is therefore selected by the current control context.
The same trace bank can yield different recalled content under different decision states, making recall addressable rather than purely similarity-based.
A recalled trace is useful only if it becomes a state the policy can act on.
We fuse the recalled trace with the control context and update a fast working state:
\begin{equation}
    w_t^\ell=f_w^\ell([u_t^\ell,r_t^\ell]),\qquad
    h_{1:T}^\ell=\mathrm{SSM}_{\mathrm{fast}}^\ell(w_{1:T}^\ell).
    \label{eq:working_state}
\end{equation}
The slow stream maintains event evidence; the fast stream maintains the policy-facing working state.
The layer then writes memory-conditioned information back into the event tokens:
\begin{equation}
    Z_t^{\ell+1}=\bar{Z}_t^\ell + f_{\mathrm{upd}}^\ell([E_t^\ell,h_t^\ell]).
    \label{eq:token_update}
\end{equation}
Stacking layers repeats event binding, trace propagation, control-indexed recall, and working-state consolidation.
The final prospective policy-facing working state is $h_t=h_t^{L}$.



\subsection{Prospective Training with Control-JEPA}
\label{sec:method_control_jepa}
Observation--action delay creates a credit-assignment problem: the value of storing an event may appear only several steps later.
Training memory only through the final action loss, therefore, provides a weak and delayed signal.
We introduce Control-JEPA to train the causal working state $h_t$ to predict future control contexts, rather than reconstruct images, past observations, or full world states.
During training, an EMA target branch encodes future embodied event tokens into target control contexts $\tilde{u}_{t+k}$, and this branch is discarded at inference.
For each horizon $k\in\mathcal{K}$, a JEPA predictor receives the causal working state $h_t$ and a horizon embedding $\eta_k$:
\begin{equation}
    \hat{u}_{t+k}=g_{\theta}\!\left([h_t,\eta_k]\right).
    \label{eq:jepa_predictor}
\end{equation}
The Control-JEPA objective is
\begin{equation}
    \mathcal{L}_{\mathrm{CJEP}}=
    \sum_{k\in\mathcal{K}}\lambda_k
    \rho\!\left(\hat{u}_{t+k},\mathrm{sg}(\tilde{u}_{t+k})\right)
    +\lambda_{\mathrm{var}}\mathcal{L}_{\mathrm{var}},
    \label{eq:control_jepa}
\end{equation}
where $\rho$ is a smooth-L1 alignment loss, $\mathrm{sg}(\cdot)$ stops gradients through the EMA target, and $\mathcal{L}_{\mathrm{var}}$ prevents representation collapse.
The horizon set spans both near-term and delayed control contexts, and horizons beyond the remaining episode length are skipped.
Because the target is the context used by a later policy step, Control-JEPA makes memory prospective while preserving causal inference.

\subsection{Memory-Conditioned Action Policy}
\label{sec:method_policy}

The final prospective working state $h_t$ conditions continuous action generation through policy tokens
\begin{equation}
    C_t = W_h h_t ,
    \label{eq:policy_context}
\end{equation}
which are consumed by a transformer action head together with the current noised action chunk.
We train the head with a clean-endpoint rectified-flow objective~\cite{vaswani2017attention,lipman2022flow,liu2023flowstraight}.
Given the ground-truth normalized action horizon $A_{t:t+H-1}$, we sample $A_0\sim\mathcal{N}(0,I)$ and $\tau\sim\mathcal{U}(0,1)$, and construct
\begin{equation}
    A_{\tau}=(1-\tau)A_0+\tau A_{t:t+H-1}.
    \label{eq:flow_interpolant}
\end{equation}
Conditioned on $C_t$, the transformer predicts the clean endpoint:
\begin{equation}
    \hat{A}_{t:t+H-1}
    =
    F_{\theta}(A_{\tau},\tau,C_t),
    \qquad
    \mathcal{L}_{\mathrm{act}}
    =
    \mathbb{E}\!\left[
    \|\hat{A}_{t:t+H-1}-A_{t:t+H-1}\|_2^2
    \right].
    \label{eq:flow_loss}
\end{equation}
At inference, an action chunk initialized from Gaussian noise is iteratively updated along the rectified-flow path using the predicted clean endpoint.
The full objective is
\begin{equation}
    \mathcal{L}
    =
    \mathcal{L}_{\mathrm{act}}
    +
    \alpha\,\mathcal{L}_{\mathrm{CJEP}},
    \label{eq:total_loss}
\end{equation}
where $\alpha$ balances imitation learning and Control-JEPA.

\section{Evaluation}

\label{sec:evaluation}
Our evaluation is organized around the claims developed in Secs.~\ref{sec:intro}--\ref{sec:method}.
The first claim is diagnostic: memory failures arise from observation--action delay. \benchname{} tests this claim by separating memory-dependent decisions from target-agnostic execution, allowing us to distinguish history-recovery errors from manipulation failures. The second claim is generalization: if control-indexed prospective memory is the right remedy, \modelname{} should improve decision success on \benchname{} and transfer to public robot memory benchmarks. The third claim is mechanistic: the gains should be explained by the three functional requirements: separability, addressability, and prospectiveness. 
We test this with complementary evidence: representation probes ask whether the learned state exhibits these properties, while ablations test whether removing the corresponding mechanisms reduces downstream success. All \benchname{} comparisons are protocol-matched, and public benchmark results follow the cited source protocols. Details are provided in Appendix.

\subsection{\benchname: Diagnostic Observation--Action Delay}
\label{sec:camo_dataset}

\begin{figure}[h]
\centering
\includegraphics[page=1,width=\textwidth,trim=0bp 5.5cm 0bp 0bp,clip]{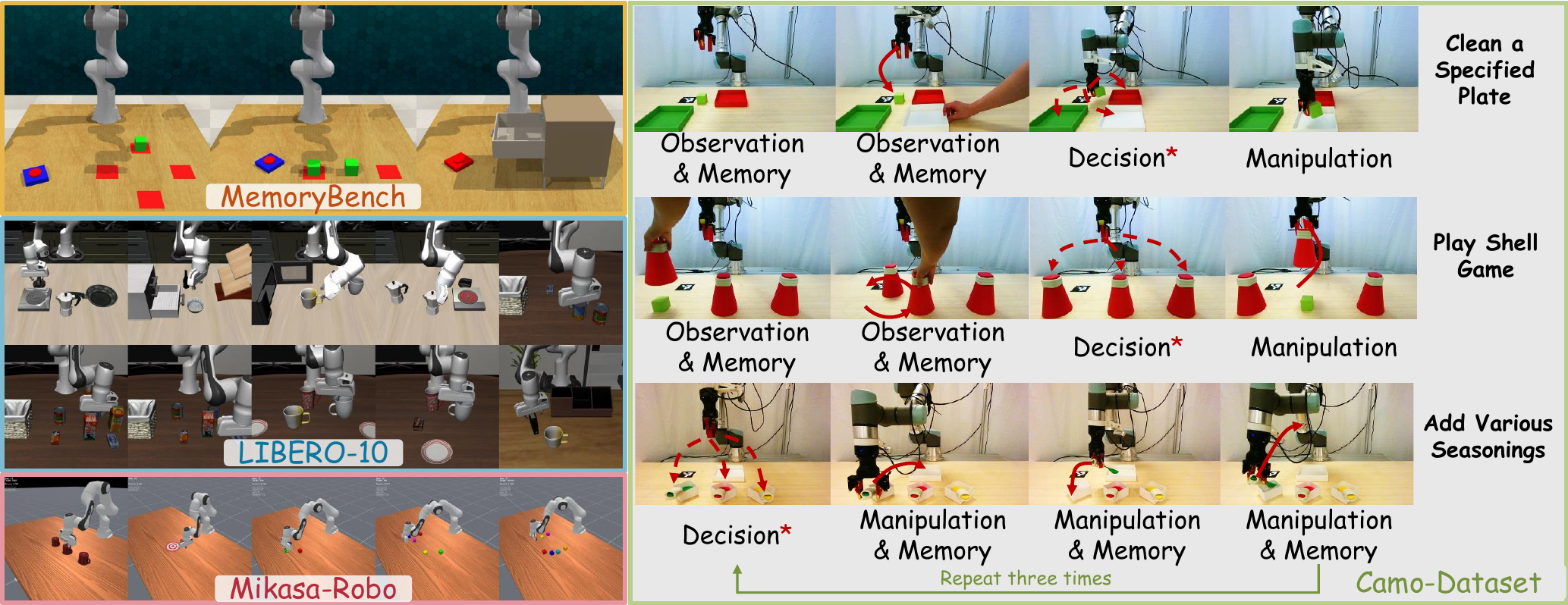}
\caption{
\textbf{Evaluation benchmarks.}
Left: public long-horizon and memory benchmarks.
Right: \benchname real-robot tasks.
The frame marked $^\ast$ denotes the non-Markovian decision stage, where the correct action depends on the earlier episode variable $z$ rather than the current observation alone.
}
\label{fig:dataset}
\end{figure}
\benchname{} is a real-robot benchmark designed to make observation--action delay measurable. Each episode contains one or more early evidence events whose control relevance is delayed until later aliased decision points. During the episode, evidence about the relevant variable $z$ may become occluded, displaced, repeated, or visually ambiguous before it is needed for action. For example, $z$ can denote the used plate in \serveplate{}, the object-containing cup in \shellgame{}, or the completed subgoals in \addpeasoning{}. At an aliased decision point, the visible scene no longer identifies the correct action:
\begin{equation}
    O_{t_d}(z_i) \approx O_{t_d}(z_j),
    \qquad
    a^\star(z_i) \neq a^\star(z_j),
    \label{eq:aliasing_condition}
\end{equation}
where $t_d$ is an annotated aliased decision time and $a^\star(z)$ is the history-correct action. Thus, \benchname{} separates memory errors from motor errors: a policy may execute a valid manipulation while selecting the wrong history-dependent target or subgoal.

We score execution and memory-dependent choice separately.
Let $M$ denote a scorable manipulation, regardless of whether the selected target or subgoal is history-correct, and let $D$ denote that the policy chooses the history-correct target or subgoal.
We report
\begin{equation}
    \mathrm{MSR}=P(M),\qquad
    \mathrm{DSR}=P(D\mid M),\qquad
    \mathrm{SR}=P(M\cap D).
    \label{eq:metrics_basic}
\end{equation}
Manipulation Success Rate (MSR) measures target-agnostic execution, Decision Success Rate (DSR) measures memory-dependent choice, and Success Rate (SR) requires both.

\subsection{Real-Robot Diagnostic Results}
\label{sec:camo_results}

\begin{table*}[!t]
\centering
\refstepcounter{table}\label{tab:experiment_result}
\setlength{\tabcolsep}{4pt}
\renewcommand{\arraystretch}{0.9}
\begin{minipage}{\textwidth}
\small\textbf{Table \thetable: Real-robot diagnostic results on \benchname.}
Task columns report DSR/SR (\%).
Avg. MSR is target-agnostic execution. Avg. columns are task-wise averages.
\end{minipage}\par
\vspace{0.35em}
\resizebox{\textwidth}{!}{
\begin{tabular}{lcccccc}
\toprule
\textbf{Method} & \textbf{\serveplate} & \textbf{\shellgame} & \textbf{\addpeasoning} & \textbf{Avg. DSR} & \textbf{Avg. MSR} & \textbf{Avg. SR} \\
& DSR/SR & DSR/SR & DSR/SR & & & \\
\midrule
\multicolumn{7}{l}{\emph{Matched imitation baselines}} \\
\DP~\cite{chi2023diffusionpolicy}
& 33.3/30.6 & 34.3/33.3 & 0.0/0.0 & 22.5 & 67.6 & 21.3 \\
ACT~\cite{zhaolearning}
& 28.0/19.4 & 35.5/30.6 & 0.0/0.0 & 21.2 & 51.8 & 16.7 \\
Flow Matching~\cite{lipman2022flow,liu2023flowstraight}
& 30.0/25.0 & 25.7/25.0 & 0.0/0.0 & 18.6 & 62.0 & 16.7 \\
\midrule
\multicolumn{7}{l}{\emph{Ours and mechanism ablations}} \\
\textbf{\modelname}
& \textbf{91.2/86.1} & \textbf{86.1/86.1} & \textbf{65.2/41.7} & \textbf{80.8} & \textbf{86.1} & \textbf{71.3} \\
\midrule
w/o memory
& 26.7/22.2 & 34.4/30.6 & 0.0/0.0 & 20.4 & 64.8 & 17.6 \\
similarity retrieval bank
& 41.4/33.3 & 28.6/22.2 & 0.0/0.0 & 23.3 & 58.3 & 18.5 \\
Vanilla Mamba memory
& 27.6/22.2 & 30.0/25.0 & 50.0/19.4 & 35.9 & 67.6 & 22.2 \\
w/o control index
& 40.7/30.6 & 45.8/30.6 & 60.0/16.7 & 48.8 & 56.5 & 26.0 \\
w/o Control-JEPA
& 82.8/66.7 & 71.0/61.1 & 61.1/30.6 & 71.6 & 72.2 & 52.8 \\
\bottomrule
\end{tabular}
}
\end{table*}

\begin{table}[!t]
\vspace{-0.5em}
\centering
\footnotesize
\setlength{\tabcolsep}{3pt}
\renewcommand{\arraystretch}{0.98}
\caption{
\textbf{Public long-horizon and memory benchmarks.}
Success rate (\%).
All entries follow the protocol of the cited source.
Benchmark details and per-task results are provided in the Appendix.
}\par
\vspace{0.35em}
\resizebox{0.98\textwidth}{!}{
\begin{tabular}{p{2.65cm} p{2.95cm} p{2.0cm} p{8.05cm}}
\toprule
\textbf{Benchmark} & \textbf{Protocol} & \textbf{\modelname} & \textbf{Published context} \\
\midrule
MemoryBench~\cite{fang2025sam2act}
& 3 task-specific policies
& \textbf{97.3 $\pm$ 4.5}
& RVT-2 54.0~\cite{goyal2024rvt2,fang2025sam2act}; SAM2Act 55.0~\cite{fang2025sam2act}; SAM2Act+ 94.3~\cite{fang2025sam2act}; ReMem-VLA 94.5$^\ast$~\cite{li2026rememvla}. \\

LIBERO-10~\cite{liu2023libero}
& 10-task mixed policy
& \textbf{87.1 $\pm$ 0.8}
& DP-T 51.0~\cite{chi2023diffusionpolicy,reuss2025efficient}; QueST 69.0~\cite{reuss2025efficient}; DP-CNN 73.0~\cite{chi2023diffusionpolicy,reuss2025efficient}; MoDE 92.0~\cite{reuss2025efficient}; OpenVLA 53.7~\cite{kim2024openvla,shi2026memoryvla}; CoT-VLA 69.0~\cite{zhao2025cot,shi2026memoryvla}; TriVLA 73.2~\cite{liu2025trivla,shi2026memoryvla}; $\pi_0$ 85.2~\cite{black2024pi_0,shi2026memoryvla}; 4D-VLA 86.5~\cite{zhang20264d,shi2026memoryvla}; MemoryVLA 93.4~\cite{shi2026memoryvla}. \\

MIKASA-Robo~\cite{cherepanov2025memory}
& 5-task mixed policy
& \textbf{75.1 $\pm$ 1.4}
& CronusVLA 18.0~\cite{li2025cronusvla,shi2026memoryvla}; SpatialVLA 21.0~\cite{qu2025spatialvla,shi2026memoryvla}; OpenVLA-OFT 28.4~\cite{kim2025openvlaoft,shi2026memoryvla}; $\pi_0$ 29.4~\cite{black2024pi_0,shi2026memoryvla}; MemoryVLA 41.2~\cite{shi2026memoryvla}; GMP 67.8~\cite{gao2026gatedmemorypolicy}. \\

MIKASA-Robo~\cite{cherepanov2025memory}
& 2 task-specific policies
& \textbf{95.6 $\pm$ 1.0}
& DP 19.5~\cite{chi2023diffusionpolicy,lei2026vpwem}; DP-PTP 15.0~\cite{lei2026vpwem}; MaIL 19.5~\cite{jia2024mail,lei2026vpwem}; DP-VPWEM 86.5~\cite{lei2026vpwem}. \\
\bottomrule
\end{tabular}
}
\label{tab:public_results}
\vspace{0.15em}
\begin{minipage}{\textwidth}
\scriptsize
$^\ast$ReMem-VLA reports a modified MemoryBench protocol.
The MIKASA-Robo 2-task row follows the specialist setting reported by VPWEM.
When a baseline score is taken from a secondary comparison table, we cite both the original method and the reporting source.
\end{minipage}
\end{table}

\renewcommand{\arraystretch}{1.0}
Table~\ref{tab:experiment_result} shows that \modelname{} resolves observation--action delay in real-robot execution: average DSR/SR improves from the strongest matched baseline, Diffusion Policy, at 22.5\%/21.3\% to 80.8\%/71.3\%.
The ablation pattern is consistent with the three design principles.
For \emph{separability}, removing memory or replacing token-grounded traces with vanilla Mamba substantially reduces performance, suggesting that delayed evidence is more useful when preserved as token-grounded traces rather than ignored or compressed into a generic sequence state.
For \emph{addressability}, similarity retrieval and removing the control index both fall well below \modelname{}, suggesting that recall is most useful when it is conditioned on the current control state rather than only on visual similarity or undirected history access.
For \emph{prospectiveness}, removing Control-JEPA retains part of the decision signal but reduces SR from 71.3\% to 52.8\%, suggesting that predicting future control context helps shape $h_t$ into a working state that is more reliable for downstream action.
These behavioral trends support the proposed memory organization, and Sec.~\ref{sec:analysis} probes the three properties directly in the learned representations.

\subsection{Public Long-Horizon and Memory Benchmarks}
\label{sec:public_results}

Table~\ref{tab:public_results} evaluates whether \modelname{} transfers beyond the diagnostic setting of \benchname{}.
On MemoryBench~\cite{fang2025sam2act}, where each task is trained separately, \modelname{} reaches 97.3\%, exceeding the original spatial-memory baselines and matching the strongest reported memory-augmented VLA result under its modified protocol.
On LIBERO-10~\cite{liu2023libero}, \modelname{} substantially improves over standard diffusion-policy and discrete-action baselines and outperforms multiple larger VLA systems.
On MIKASA-Robo~\cite{cherepanov2025memory}, the strongest public test of non-Markovian task memory in our evaluation, \modelname{} achieves the best reported result in both settings: it improves the 5-task mixed-policy average from 67.8\% to 75.1\%, and the 2-task specialist result from 86.5\% to 95.6\%. Detailed benchmark protocols, per-task scores, and source-specific comparisons are provided in the Appendix.

\subsection{Mechanistic Probes}

\begin{figure}[!htbp]
\centering
\includegraphics[page=1,width=\textwidth,trim=57bp 121bp 100bp 91bp,clip]{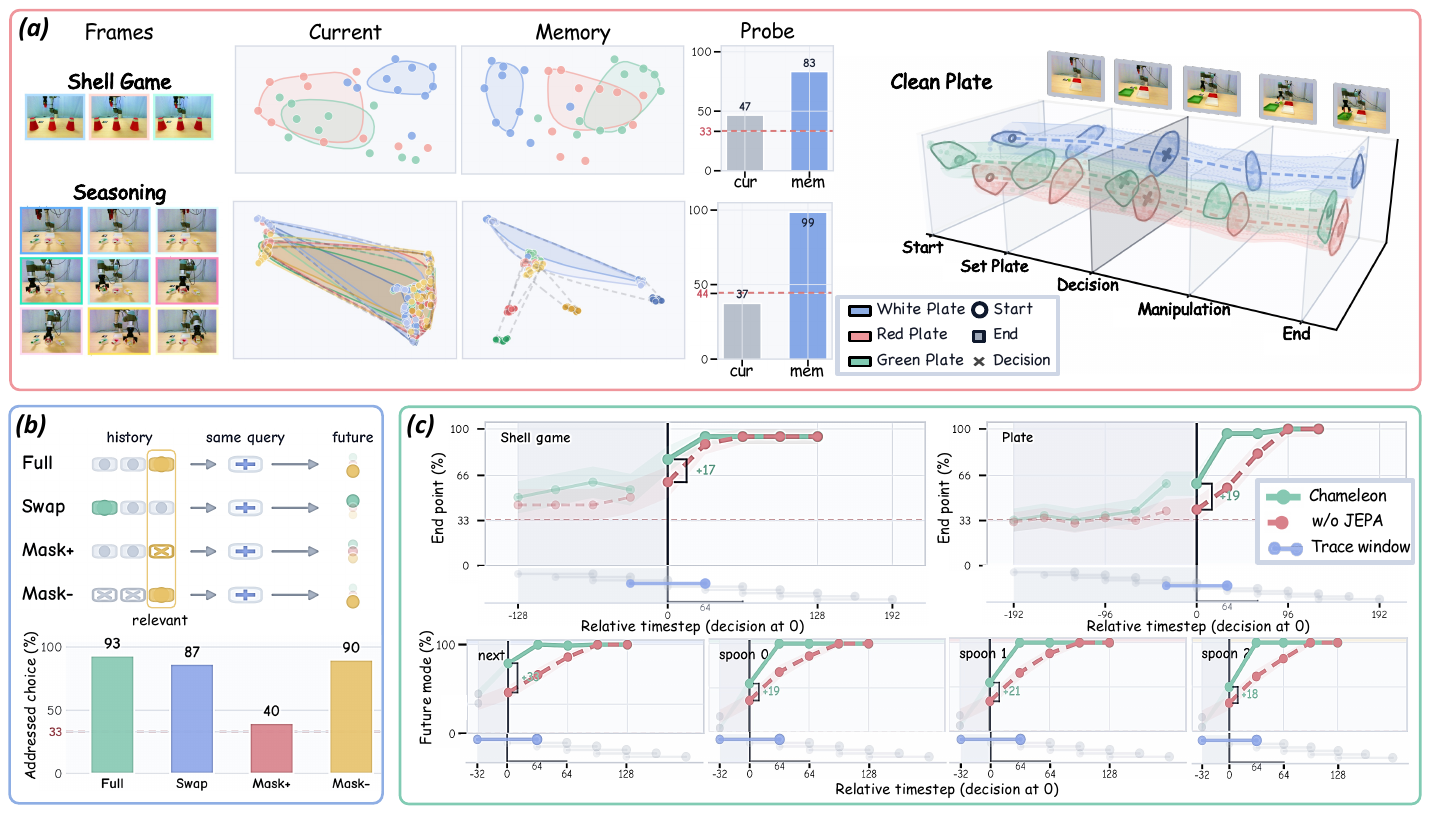}
\caption{
\textbf{Mechanistic probes on \benchname.}
(a) \emph{Separability}: hidden-variable decoding from current tokens and memory states.
(b) \emph{Addressability}: counterfactual trace edits under a fixed control query.
(c) \emph{Prospectiveness}: future-control decoding from $h_t$ with and without Control-JEPA.
}
\label{fig:analysis-completion}
\end{figure}

\label{sec:analysis}

\vspace{-0.5em}
\paragraph{Separability.}
Figure~\ref{fig:analysis-completion}(a) compares hidden-variable probes on \emph{current} and \emph{memory} representations at aliased decision frames. These correspond to current event tokens $Z_t^0$ and the final policy-facing memory state $h_t^L$.
Probes on memory representations are more accurate than probes on current representations on \shellgame{} (83.3\% vs. 46.7\%) and \addpeasoning{} (98.5\% vs. 37.4\%).
The \serveplate{} visualization provides a qualitative example where memory trajectories remain organized by earlier plate identity after visual aliasing.
\vspace{-0.5em}
\paragraph{Addressability.}
Figure~\ref{fig:analysis-completion}(b) tests whether the control query $u_t^\ell$ addresses the relevant trace in $E_t^\ell$.
In \addpeasoning{}, we fix $u_t^\ell$ and intervene on $E_t^\ell$: Full keeps the original history, Swap replaces the relevant trace, Mask+ removes it, and Mask- removes irrelevant traces.
We report counterfactual choice accuracy, i.e., whether the selected subgoal follows the active relevant trace.
The pattern Full 93\%, Swap 87\%, Mask+ 40\%, and Mask- 90\% shows that recall is selected by control relevance rather than temporal proximity or visual similarity.


\vspace{-0.5em}
\paragraph{Prospectiveness.}
Figure~\ref{fig:analysis-completion}(c) asks whether $h_t$ already contains future action information.
We decode the later target endpoint in reaching tasks and the next subgoal mode in sequential tasks.
Compared with w/o Control-JEPA, \modelname{} exposes this information earlier and more reliably, showing that Control-JEPA makes memory more prospective and action-ready.

\section{Conclusion and Limitations}
\label{sec:conclusion}

\vspace{-0.5em}

\paragraph{Conclusion.}
We introduced observation--action delay as a control-level memory failure: by the time the robot must act, the current observation no longer contains the information needed to execute correctly.
Chameleon addresses this failure with control-indexed prospective memory, keeping histories separable, recalling the control-relevant trace, and consolidating recall into an action-ready state.
Across diagnostic real-robot tasks, public memory benchmarks, ablations, and probes, Chameleon achieves state-of-the-art performance among same-size policies, surpasses multiple larger VLA baselines under reported protocols, and realizes the intended properties of separability, addressability, and prospectiveness.
These results suggest that robot memory should be designed not merely to store more of the past, but to make the right past actionable at the right moment.

\vspace{-0.5em}

\paragraph{Limitations and Future Directions.}
This work studies control-indexed prospective memory within episode-level imitation policies.
A natural next question is how the same memory organization scales across substantially different embodiments, sensor layouts, and task families.
One promising direction is to make control-indexed memory a reusable module for foundation-scale robot policies, supporting cross-task, cross-embodiment, and cross-environment generalization.
Another is to couple prospective memory with active perception, allowing robots to acquire or refresh evidence before it becomes action-critical.

\clearpage
\bibliographystyle{plainnat}
\bibliography{paper}

\begin{thebibliography}{53}
\providecommand{\natexlab}[1]{#1}
\providecommand{\url}[1]{\texttt{#1}}
\expandafter\ifx\csname urlstyle\endcsname\relax
  \providecommand{\doi}[1]{doi: #1}\else
  \providecommand{\doi}{doi: \begingroup \urlstyle{rm}\Url}\fi

\bibitem[Allen and Fortin(2013)]{allen2013evolution}
Timothy~A Allen and Norbert~J Fortin.
\newblock The evolution of episodic memory.
\newblock \emph{Proceedings of the National Academy of Sciences}, 110\penalty0 (supplement\_2):\penalty0 10379--10386, 2013.

\bibitem[Anwar et~al.(2025)Anwar, Welsh, Biswas, Pouya, and Chang]{anwar2025remembr}
Abrar Anwar, John Welsh, Joydeep Biswas, Soha Pouya, and Yan Chang.
\newblock Remembr: Building and reasoning over long-horizon spatio-temporal memory for robot navigation.
\newblock In \emph{2025 IEEE International Conference on Robotics and Automation (ICRA)}, pages 2838--2845. IEEE, 2025.

\bibitem[Assran et~al.(2023)Assran, Duval, Misra, Bojanowski, Vincent, Rabbat, LeCun, and Ballas]{assran2023selfsupervised}
Mahmoud Assran, Quentin Duval, Ishan Misra, Piotr Bojanowski, Pascal Vincent, Michael Rabbat, Yann LeCun, and Nicolas Ballas.
\newblock Self-supervised learning from images with a joint-embedding predictive architecture.
\newblock In \emph{Proceedings of the IEEE/CVF conference on computer vision and pattern recognition}, pages 15619--15629, 2023.

\bibitem[Bakker et~al.(2008)Bakker, Kirwan, Miller, and Stark]{bakker2008pattern}
Arnold Bakker, C~Brock Kirwan, Michael Miller, and Craig~EL Stark.
\newblock Pattern separation in the human hippocampal ca3 and dentate gyrus.
\newblock \emph{science}, 319\penalty0 (5870):\penalty0 1640--1642, 2008.

\bibitem[Black et~al.(2024)Black, Brown, Driess, Esmail, Equi, Finn, Fusai, Groom, Hausman, Ichter, et~al.]{black2024pi_0}
Kevin Black, Noah Brown, Danny Driess, Adnan Esmail, Michael Equi, Chelsea Finn, Niccolo Fusai, Lachy Groom, Karol Hausman, Brian Ichter, et~al.
\newblock $\pi\_0 $: A vision-language-action flow model for general robot control.
\newblock \emph{arXiv preprint arXiv:2410.24164}, 2024.

\bibitem[Borgeaud et~al.(2022)Borgeaud, Mensch, Hoffmann, Cai, Rutherford, Millican, Van Den~Driessche, Lespiau, Damoc, Clark, et~al.]{borgeaud2022improving}
Sebastian Borgeaud, Arthur Mensch, Jordan Hoffmann, Trevor Cai, Eliza Rutherford, Katie Millican, George~Bm Van Den~Driessche, Jean-Baptiste Lespiau, Bogdan Damoc, Aidan Clark, et~al.
\newblock Improving language models by retrieving from trillions of tokens.
\newblock In \emph{International conference on machine learning}, pages 2206--2240. PMLR, 2022.

\bibitem[Cherepanov et~al.(2026)Cherepanov, Kachaev, Kovalev, and Panov]{cherepanov2025memory}
Egor Cherepanov, Nikita Kachaev, Alexey Kovalev, and Aleksandr Panov.
\newblock Memory, benchmark \& robots: A benchmark for solving complex tasks with reinforcement learning.
\newblock In \emph{The Fourteenth International Conference on Learning Representations}, 2026.
\newblock URL \url{https://openreview.net/forum?id=9cLPurIZMj}.

\bibitem[Chi et~al.(2023)Chi, Feng, Du, Xu, Cousineau, Burchfiel, and Song]{chi2023diffusionpolicy}
Cheng Chi, Siyuan Feng, Yilun Du, Zhenjia Xu, Eric Cousineau, Benjamin Burchfiel, and Shuran Song.
\newblock Diffusion policy: Visuomotor policy learning via action diffusion.
\newblock In \emph{Proceedings of Robotics: Science and Systems (RSS)}, 2023.

\bibitem[Chung et~al.(2026)Chung, Hanyu, Nguyen, Le, Bumgarner, Nguyen, Vo, Yamazaki, Rainwater, Kieu, et~al.]{chung2025rethinking}
Nhat Chung, Taisei Hanyu, Toan Nguyen, Huy Le, Frederick Bumgarner, Duy Minh~Ho Nguyen, Khoa Vo, Kashu Yamazaki, Chase Rainwater, Tung Kieu, et~al.
\newblock Rethinking progression of memory state in robotic manipulation: An object-centric perspective.
\newblock In \emph{Proceedings of the AAAI Conference on Artificial Intelligence}, volume~40, pages 3407--3415, 2026.

\bibitem[Dao and Gu(2024)]{dao2024mamba2}
Tri Dao and Albert Gu.
\newblock Transformers are {SSM}s: Generalized models and efficient algorithms through structured state space duality.
\newblock In \emph{International Conference on Machine Learning (ICML)}, 2024.

\bibitem[Fang et~al.(2025)Fang, Grotz, Pumacay, Wang, Fox, Krishna, and Duan]{fang2025sam2act}
Haoquan Fang, Markus Grotz, Wilbert Pumacay, Yi~Ru Wang, Dieter Fox, Ranjay Krishna, and Jiafei Duan.
\newblock Sam2act: Integrating visual foundation model with a memory architecture for robotic manipulation.
\newblock \emph{arXiv preprint arXiv:2501.18564}, 2025.

\bibitem[Gao et~al.(2026)Gao, Liu, Li, and Song]{gao2026gatedmemorypolicy}
Yihuai Gao, Jinyun Liu, Shuang Li, and Shuran Song.
\newblock Gated memory policy, 2026.
\newblock URL \url{https://arxiv.org/abs/2604.18933}.

\bibitem[Goyal et~al.(2024)Goyal, Blukis, Xu, Guo, Chao, and Fox]{goyal2024rvt2}
Ankit Goyal, Valts Blukis, Jie Xu, Yijie Guo, Yu-Wei Chao, and Dieter Fox.
\newblock Rvt-2: Learning precise manipulation from few demonstrations.
\newblock \emph{arXiv preprint arXiv:2406.08545}, 2024.

\bibitem[Gu and Dao(2023)]{gu2023mamba}
Albert Gu and Tri Dao.
\newblock Mamba: Linear-time sequence modeling with selective state spaces.
\newblock \emph{arXiv preprint arXiv:2312.00752}, 2023.

\bibitem[Gu et~al.(2022)Gu, Goel, and R\'e]{gu2022efficiently}
Albert Gu, Karan Goel, and Christopher R\'e.
\newblock Efficiently modeling long sequences with structured state spaces.
\newblock In \emph{The International Conference on Learning Representations ({ICLR})}, 2022.

\bibitem[Hawthorne et~al.(2022)Hawthorne, Jaegle, Cangea, Borgeaud, Nash, Malinowski, Dieleman, Vinyals, Botvinick, Simon, et~al.]{hawthorne2022general}
Curtis Hawthorne, Andrew Jaegle, C{\u{a}}t{\u{a}}lina Cangea, Sebastian Borgeaud, Charlie Nash, Mateusz Malinowski, Sander Dieleman, Oriol Vinyals, Matthew Botvinick, Ian Simon, et~al.
\newblock General-purpose, long-context autoregressive modeling with perceiver ar.
\newblock In \emph{International Conference on Machine Learning}, pages 8535--8558. PMLR, 2022.

\bibitem[Hu et~al.(2026)Hu, Cui, Lu, Yang, Ye, Zhao, Chen, Lan, and Ren]{hu2026echo}
Yanbin Hu, Jin Cui, Jiayi Lu, Ruixuan Yang, Jun Ye, Boran Zhao, Xingyu Chen, Xuguang Lan, and Pengju Ren.
\newblock Echo: Continuous hierarchical memory for vision-language-action models, 2026.
\newblock URL \url{https://arxiv.org/abs/2605.10993}.

\bibitem[Jia et~al.(2024)Jia, Wang, Donat, Xing, Li, Zhou, Celik, Blessing, Lioutikov, and Neumann]{jia2024mail}
Xiaogang Jia, Qian Wang, Atalay Donat, Bowen Xing, Ge~Li, Hongyi Zhou, Onur Celik, Denis Blessing, Rudolf Lioutikov, and Gerhard Neumann.
\newblock Mail: Improving imitation learning with selective state space models.
\newblock In \emph{8th Annual Conference on Robot Learning}, 2024.

\bibitem[Karpukhin et~al.(2020)Karpukhin, Oguz, Min, Lewis, Wu, Edunov, Chen, and Yih]{karpukhin2020dense}
Vladimir Karpukhin, Barlas Oguz, Sewon Min, Patrick Lewis, Ledell Wu, Sergey Edunov, Danqi Chen, and Wen-tau Yih.
\newblock Dense passage retrieval for open-domain question answering.
\newblock In \emph{Proceedings of the 2020 conference on empirical methods in natural language processing (EMNLP)}, pages 6769--6781, 2020.

\bibitem[Kim et~al.(2024)Kim, Pertsch, Karamcheti, Xiao, Balakrishna, Nair, Rafailov, Foster, Lam, Sanketi, et~al.]{kim2024openvla}
Moo~Jin Kim, Karl Pertsch, Siddharth Karamcheti, Ted Xiao, Ashwin Balakrishna, Suraj Nair, Rafael Rafailov, Ethan Foster, Grace Lam, Pannag Sanketi, et~al.
\newblock Openvla: An open-source vision-language-action model.
\newblock \emph{arXiv preprint arXiv:2406.09246}, 2024.

\bibitem[Kim et~al.(2025)Kim, Finn, and Liang]{kim2025openvlaoft}
Moo~Jin Kim, Chelsea Finn, and Percy Liang.
\newblock Fine-tuning vision-language-action models: Optimizing speed and success.
\newblock \emph{arXiv preprint arXiv:2502.19645}, 2025.

\bibitem[Lei et~al.(2026)Lei, Liang, Zhang, and Luo]{lei2026vpwem}
Yuheng Lei, Zhixuan Liang, Hongyuan Zhang, and Ping Luo.
\newblock Vpwem: Non-markovian visuomotor policy with working and episodic memory.
\newblock \emph{arXiv preprint arXiv:2603.04910}, 2026.

\bibitem[Lewis et~al.(2020)Lewis, Perez, Piktus, Petroni, Karpukhin, Goyal, K{\"u}ttler, Lewis, Yih, Rockt{\"a}schel, et~al.]{lewis2020retrieval}
Patrick Lewis, Ethan Perez, Aleksandra Piktus, Fabio Petroni, Vladimir Karpukhin, Naman Goyal, Heinrich K{\"u}ttler, Mike Lewis, Wen-tau Yih, Tim Rockt{\"a}schel, et~al.
\newblock Retrieval-augmented generation for knowledge-intensive nlp tasks.
\newblock \emph{Advances in neural information processing systems}, 33:\penalty0 9459--9474, 2020.

\bibitem[Li et~al.(2026)Li, Shen, Chen, Yang, Wang, Shi, Bing, Liu, and Knoll]{li2026rememvla}
Hang Li, Fengyi Shen, Dong Chen, Liudi Yang, Xudong Wang, Jinkui Shi, Zhenshan Bing, Ziyuan Liu, and Alois Knoll.
\newblock Remem-vla: Empowering vision-language-action model with memory via dual-level recurrent queries, 2026.
\newblock URL \url{https://arxiv.org/abs/2603.12942}.

\bibitem[Li et~al.(2025{\natexlab{a}})Li, Yang, Chen, Tian, Yang, Chen, Wang, Wang, Zhao, Lin, et~al.]{li2025cronusvla}
Hao Li, Shuai Yang, Yilun Chen, Yang Tian, Xiaoda Yang, Xinyi Chen, Hanqing Wang, Tai Wang, Feng Zhao, Dahua Lin, et~al.
\newblock Cronusvla: Transferring latent motion across time for multi-frame prediction in manipulation.
\newblock \emph{arXiv e-prints}, pages arXiv--2506, 2025{\natexlab{a}}.

\bibitem[Li et~al.(2025{\natexlab{b}})Li, Guo, Wu, Wang, Deng, Weng, Tan, and Wang]{li2025mapvla}
Runhao Li, Wenkai Guo, Zhenyu Wu, Changyuan Wang, Haoyuan Deng, Zhenyu Weng, Yap-Peng Tan, and Ziwei Wang.
\newblock Map-vla: Memory-augmented prompting for vision-language-action model in robotic manipulation, 2025{\natexlab{b}}.
\newblock URL \url{https://arxiv.org/abs/2511.09516}.

\bibitem[Lin et~al.(2025)Lin, Liang, Lin, Jingzhi, Jiao, Li, Ma, Liu, Zhao, Zhuang, et~al.]{lin2025echovla}
Min Lin, Xiwen Liang, Bingqian Lin, Liu Jingzhi, Zijian Jiao, Kehan Li, Yuhan Ma, Yuecheng Liu, Shen Zhao, Yuzheng Zhuang, et~al.
\newblock Echovla: Robotic vision-language-action model with synergistic declarative memory for mobile manipulation.
\newblock \emph{arXiv preprint arXiv:2511.18112}, 2025.

\bibitem[Lipman et~al.(2022)Lipman, Chen, Ben-Hamu, Nickel, and Le]{lipman2022flow}
Yaron Lipman, Ricky~TQ Chen, Heli Ben-Hamu, Maximilian Nickel, and Matt Le.
\newblock Flow matching for generative modeling.
\newblock \emph{arXiv preprint arXiv:2210.02747}, 2022.

\bibitem[Liu et~al.(2023)Liu, Zhu, Gao, Feng, Liu, Zhu, and Stone]{liu2023libero}
Bo~Liu, Yifeng Zhu, Chongkai Gao, Yihao Feng, Qiang Liu, Yuke Zhu, and Peter Stone.
\newblock Libero: Benchmarking knowledge transfer for lifelong robot learning.
\newblock \emph{Advances in Neural Information Processing Systems}, 36:\penalty0 44776--44791, 2023.

\bibitem[Liu et~al.(2022)Liu, Gong, and Liu]{liu2023flowstraight}
Xingchao Liu, Chengyue Gong, and Qiang Liu.
\newblock Flow straight and fast: Learning to generate and transfer data with rectified flow.
\newblock \emph{arXiv preprint arXiv:2209.03003}, 2022.

\bibitem[Liu et~al.(2024)Liu, Song, Jiang, Chen, Luo, Li, and Lin]{liu2024meia}
Yang Liu, Xinshuai Song, Kaixuan Jiang, Weixing Chen, Jingzhou Luo, Guanbin Li, and Liang Lin.
\newblock Meia: Multimodal embodied perception and interaction in unknown environments.
\newblock \emph{arXiv preprint arXiv:2402.00290}, 2024.

\bibitem[Liu et~al.(2025)Liu, Gu, Zheng, Fu, Xue, and Jiang]{liu2025trivla}
Zhenyang Liu, Yongchong Gu, Sixiao Zheng, Yanwei Fu, Xiangyang Xue, and Yu-Gang Jiang.
\newblock Trivla: A triple-system-based unified vision-language-action model with episodic world modeling for general robot control.
\newblock \emph{arXiv preprint arXiv:2507.01424}, 2025.

\bibitem[Mon-Williams et~al.(2025)Mon-Williams, Li, Long, Du, and Lucas]{mon2025embodied}
Ruaridh Mon-Williams, Gen Li, Ran Long, Wenqian Du, and Christopher~G Lucas.
\newblock Embodied large language models enable robots to complete complex tasks in unpredictable environments.
\newblock \emph{Nature Machine Intelligence}, 7\penalty0 (4):\penalty0 592--601, 2025.

\bibitem[Parisotto et~al.(2020)Parisotto, Song, Rae, Pascanu, Gulcehre, Jayakumar, Jaderberg, Kaufman, Clark, Noury, et~al.]{parisotto2020stabilizing}
Emilio Parisotto, Francis Song, Jack Rae, Razvan Pascanu, Caglar Gulcehre, Siddhant Jayakumar, Max Jaderberg, Raphael~Lopez Kaufman, Aidan Clark, Seb Noury, et~al.
\newblock Stabilizing transformers for reinforcement learning.
\newblock In \emph{International conference on machine learning}, pages 7487--7498. PMLR, 2020.

\bibitem[Park et~al.(2023)Park, O'Brien, Cai, Morris, Liang, and Bernstein]{park2023generative}
Joon~Sung Park, Joseph O'Brien, Carrie~Jun Cai, Meredith~Ringel Morris, Percy Liang, and Michael~S Bernstein.
\newblock Generative agents: Interactive simulacra of human behavior.
\newblock In \emph{Proceedings of the 36th annual acm symposium on user interface software and technology}, pages 1--22, 2023.

\bibitem[Qian et~al.(2026)Qian, He, Shi, Xiao, and Jiang]{qian2026escape}
Jingjing Qian, Zeyuan He, Chen Shi, Lei Xiao, and Li~Jiang.
\newblock Escape: Episodic spatial memory and adaptive execution policy for long-horizon mobile manipulation, 2026.
\newblock URL \url{https://arxiv.org/abs/2604.13633}.

\bibitem[Qu et~al.(2025)Qu, Song, Chen, Yao, Ye, Ding, Wang, Gu, Zhao, Wang, et~al.]{qu2025spatialvla}
Delin Qu, Haoming Song, Qizhi Chen, Yuanqi Yao, Xinyi Ye, Yan Ding, Zhigang Wang, JiaYuan Gu, Bin Zhao, Dong Wang, et~al.
\newblock Spatialvla: Exploring spatial representations for visual-language-action model.
\newblock \emph{arXiv preprint arXiv:2501.15830}, 2025.

\bibitem[Reuss et~al.(2025)Reuss, Pari, Agrawal, and Lioutikov]{reuss2025efficient}
Moritz Reuss, Jyothish Pari, Pulkit Agrawal, and Rudolf Lioutikov.
\newblock Efficient diffusion transformer policies with mixture of expert denoisers for multitask learning.
\newblock In \emph{The Thirteenth International Conference on Learning Representations}, 2025.
\newblock URL \url{https://openreview.net/forum?id=nDmwloEl3N}.

\bibitem[Sanh et~al.(2020)Sanh, Debut, Chaumond, and Wolf]{sanh2019distilbert}
Victor Sanh, Lysandre Debut, Julien Chaumond, and Thomas Wolf.
\newblock Distilbert, a distilled version of bert: smaller, faster, cheaper and lighter, 2020.
\newblock URL \url{https://arxiv.org/abs/1910.01108}.

\bibitem[Shi et~al.(2026)Shi, Xie, Liu, Sun, Liu, Wang, Zhou, Fan, Zhang, and Huang]{shi2026memoryvla}
Hao Shi, Bin Xie, Yingfei Liu, Lin Sun, Fengrong Liu, Tiancai Wang, Erjin Zhou, Haoqiang Fan, Xiangyu Zhang, and Gao Huang.
\newblock Memory{VLA}: Perceptual-cognitive memory in vision-language-action models for robotic manipulation.
\newblock In \emph{The Fourteenth International Conference on Learning Representations}, 2026.
\newblock URL \url{https://openreview.net/forum?id=54U3XHf7qq}.

\bibitem[Sridhar et~al.(2026)Sridhar, Pan, Sharma, and Finn]{sridhar2025memer}
Ajay Sridhar, Jennifer Pan, Satvik Sharma, and Chelsea Finn.
\newblock Scaling up memory for robotic control via experience retrieval.
\newblock In \emph{The Fourteenth International Conference on Learning Representations}, 2026.
\newblock URL \url{https://openreview.net/forum?id=1dH4ARGdwD}.

\bibitem[Torne et~al.(2026)Torne, Pertsch, Walke, Vedder, Nair, Ichter, Ren, Wang, Tang, Stachowicz, Dhabalia, Equi, Vuong, Springenberg, Levine, Finn, and Driess]{torne2026vlas}
Marcel Torne, Karl Pertsch, Homer Walke, Kyle Vedder, Suraj Nair, Brian Ichter, Allen~Z. Ren, Haohuan Wang, Jiaming Tang, Kyle Stachowicz, Karan Dhabalia, Michael Equi, Quan Vuong, Jost~Tobias Springenberg, Sergey Levine, Chelsea Finn, and Danny Driess.
\newblock Mem: Multi-scale embodied memory for vision language action models, 2026.
\newblock URL \url{https://arxiv.org/abs/2603.03596}.

\bibitem[Vaswani et~al.(2017)Vaswani, Shazeer, Parmar, Uszkoreit, Jones, Gomez, Kaiser, and Polosukhin]{vaswani2017attention}
Ashish Vaswani, Noam Shazeer, Niki Parmar, Jakob Uszkoreit, Llion Jones, Aidan~N Gomez, {\L}ukasz Kaiser, and Illia Polosukhin.
\newblock Attention is all you need.
\newblock \emph{Advances in neural information processing systems}, 30, 2017.

\bibitem[Wang et~al.(2025)Wang, Yu, Zhao, Sun, Hou, Liang, Hu, Han, and Gan]{wang2025karma}
Zixuan Wang, Bo~Yu, Junzhe Zhao, Wenhao Sun, Sai Hou, Shuai Liang, Xing Hu, Yinhe Han, and Yiming Gan.
\newblock Karma: Augmenting embodied ai agents with long-and-short term memory systems.
\newblock In \emph{2025 IEEE International Conference on Robotics and Automation (ICRA)}, pages 1--8. IEEE, 2025.

\bibitem[Worsfold et~al.(2025)Worsfold, Clayton, and Cheke]{worsfold2025revisiting}
Ella Worsfold, Nicola~S Clayton, and Lucy~G Cheke.
\newblock Revisiting episodic-like memory in scrub jays: Is there more we can still learn from what--where--when caching behaviour?
\newblock \emph{Learning \& Behavior}, 53\penalty0 (1):\penalty0 65--79, 2025.

\bibitem[Xie et~al.(2024)Xie, Min, Ji, Yang, Zhang, Xu, Bajaj, Salakhutdinov, Johnson-Roberson, and Bisk]{xie2024embodied}
Quanting Xie, So~Yeon Min, Pengliang Ji, Yue Yang, Tianyi Zhang, Kedi Xu, Aarav Bajaj, Ruslan Salakhutdinov, Matthew Johnson-Roberson, and Yonatan Bisk.
\newblock Embodied-rag: General non-parametric embodied memory for retrieval and generation.
\newblock \emph{arXiv preprint arXiv:2409.18313}, 2024.

\bibitem[Yao et~al.(2022)Yao, Zhao, Yu, Du, Shafran, Narasimhan, and Cao]{yao2022react}
Shunyu Yao, Jeffrey Zhao, Dian Yu, Nan Du, Izhak Shafran, Karthik Narasimhan, and Yuan Cao.
\newblock React: Synergizing reasoning and acting in language models.
\newblock \emph{arXiv preprint arXiv:2210.03629}, 2022.

\bibitem[Zeng et~al.(2026)Zeng, Ding, Yang, and Li]{zeng2026helm}
Zijian Zeng, Fei Ding, Huiming Yang, and Xianwei Li.
\newblock Helm: Harness-enhanced long-horizon memory for vision-language-action manipulation.
\newblock \emph{arXiv preprint arXiv:2604.18791}, 2026.

\bibitem[Zhang et~al.(2026)Zhang, Chen, Xu, Huang, Zhou, Yuan, Cai, Huang, Quan, Xu, et~al.]{zhang20264d}
Jiahui Zhang, Yurui Chen, Yueming Xu, Ze~Huang, Yanpeng Zhou, Yu-Jie Yuan, Xinyue Cai, Guowei Huang, Xingyue Quan, Hang Xu, et~al.
\newblock 4d-vla: Spatiotemporal vision-language-action pretraining with cross-scene calibration.
\newblock \emph{Advances in Neural Information Processing Systems}, 38:\penalty0 33914--33937, 2026.

\bibitem[Zhao et~al.(2025)Zhao, Lu, Kim, Fu, Zhang, Wu, Li, Ma, Han, Finn, et~al.]{zhao2025cot}
Qingqing Zhao, Yao Lu, Moo~Jin Kim, Zipeng Fu, Zhuoyang Zhang, Yecheng Wu, Zhaoshuo Li, Qianli Ma, Song Han, Chelsea Finn, et~al.
\newblock Cot-vla: Visual chain-of-thought reasoning for vision-language-action models.
\newblock In \emph{Proceedings of the Computer Vision and Pattern Recognition Conference}, pages 1702--1713, 2025.

\bibitem[Zhao et~al.(2023)Zhao, Kumar, Levine, and Finn]{zhaolearning}
Tony~Z Zhao, Vikash Kumar, Sergey Levine, and Chelsea Finn.
\newblock Learning fine-grained bimanual manipulation with low-cost hardware.
\newblock \emph{arXiv preprint arXiv:2304.13705}, 2023.

\bibitem[Zheng et~al.(2025)Zheng, Wolf, Ranganath, O'Reilly, and McKee]{zheng2025flexible}
Yicong Zheng, Nora Wolf, Charan Ranganath, Randall~C O'Reilly, and Kevin~L McKee.
\newblock Flexible prefrontal control over hippocampal episodic memory for goal-directed generalization.
\newblock \emph{arXiv preprint arXiv:2503.02303}, 2025.

\bibitem[Zhu et~al.(2024)Zhu, Ou, Mou, and Tang]{zhu2024retrieval}
Yichen Zhu, Zhicai Ou, Xiaofeng Mou, and Jian Tang.
\newblock Retrieval-augmented embodied agents.
\newblock In \emph{Proceedings of the IEEE/CVF Conference on Computer Vision and Pattern Recognition}, pages 17985--17995, 2024.

\end{thebibliography}

\clearpage
\appendix
\section*{Appendices}
\addcontentsline{toc}{section}{Appendices}
Within this supplementary material, we elaborate on the following aspects:
\vspace{1em}
\begin{itemize}
  \item \textbf{Appendix A:} Method Details and Derivations
    \begin{itemize}
      \item \textbf{A.1:} Notation and Dimension Bookkeeping
      \item \textbf{A.2:} Control-Indexed Memory Cell and Control-JEPA
    \end{itemize}

  \item \textbf{Appendix B:} Experimental Setup
    \begin{itemize}
      \item \textbf{B.1:} Simulation Benchmarks
      \item \textbf{B.2:} Real-Robot Camo-Dataset
      \item \textbf{B.3:} Evaluation Metrics
    \end{itemize}

  \item \textbf{Appendix C:} Implementation Details of Chameleon
    \begin{itemize}
      \item \textbf{C.1:} Parameter Budget
      \item \textbf{C.2:} Hyperparameters
    \end{itemize}

  \item \textbf{Appendix D:} Per-Task Results on Public Benchmarks

  \item \textbf{Appendix E:} Full Ablations with Statistics

  \item \textbf{Appendix F:} Mechanistic Probe Details
    \begin{itemize}
      \item \textbf{F.1:} Separability Probe
      \item \textbf{F.2:} Addressability Probe
      \item \textbf{F.3:} Prospectiveness Probe
    \end{itemize}
\end{itemize}
\clearpage

\setcounter{table}{0}
\setcounter{figure}{0}
\renewcommand{\thetable}{S\arabic{table}}
\renewcommand{\thefigure}{S\arabic{figure}}



\section{Method Details and Derivations}
\label{app:method}

\subsection{Notation and Dimension Bookkeeping}
\label{app:dims}

Table~\ref{tab:notation} lists every symbol used in Sec.~3 together with its shape. Throughout,
$d$ is the shared token dimension, $N$ is the number of embodied event tokens per timestep, and
$L$ is the number of memory layers. The token count $N$ is fixed within a benchmark but differs
across benchmarks because the number of camera views $V$ differs; concretely,
\begin{equation}
N \;=\; \sum_{v=1}^{V} N_v \;+\; 2,
\end{equation}
where $N_v$ is the patch-token count of view $v$ and the $+2$ accounts for the single
proprioception token $p_t$ and the single language (or null-instruction) token $q_t$
(Eq.~4). Per-benchmark values of $V$, $N_v$, and $N$ are given in Table~\ref{tab:hparams}.

\begin{table}[h]
\centering
\setlength{\tabcolsep}{8pt}
\caption{\textbf{Notation and tensor shapes.} Shapes are per timestep $t$ and per layer $\ell$
unless noted. $B$ denotes batch size and is omitted from the shape column.}
\begin{tabular}{l l l}
\toprule
\textbf{Symbol} & \textbf{Shape} & \textbf{Description} \\
\midrule
$d$              & scalar          & token / hidden dimension \\
$V$              & scalar          & number of camera views \\
$N_v$            & scalar          & patch tokens from view $v$ \\
$N$              & scalar          & total event tokens, $N=\sum_v N_v+2$ \\
$L$              & scalar          & number of memory layers \\
$H$              & scalar          & action prediction horizon \\
$X^v_t$          & $N_v \times d$  & patch tokens of view $v$ (Eq.~\ref{eq:visual_tokens}) \\
$p_t,\,q_t$      & $d$             & proprio / language tokens (Eq.~\ref{eq:prop_lang_tokens}) \\
$Z^\ell_t$       & $N \times d$    & event tokens entering layer $\ell$ \\
$\bar{Z}^\ell_t$ & $N \times d$    & event tokens after binding (Eq.~\ref{eq:event_mixer}) \\
$E^\ell_t$       & $N \times d$    & token-grounded traces, $\{e^\ell_{t,i}\}_{i=1}^{N}$ (Eq.~\ref{eq:slow_episodic}) \\
$c^\ell_t$       & $d$             & control index (Eq.~\ref{eq:control_seed}) \\
$u^\ell_t$       & $d$             & control context (Eq.~\ref{eq:control_context}) \\
$r^\ell_t$       & $d$             & recalled trace (Eq.~\ref{eq:episodic_recall}) \\
$w^\ell_t$       & $d$             & fused working input (Eq.~\ref{eq:working_state}) \\
$h^\ell_t$       & $d$             & working state at layer $\ell$ (Eq.~\ref{eq:working_state}) \\
$h_t=h^L_t$      & $d$             & final policy-facing working state \\
$C_t$            & $n_c \times d$ & policy tokens (Eq.~\ref{eq:policy_context}) \\
$\hat{A}_{t:t+H-1}$ & $H \times d_a$ & predicted action horizon \\
$\tilde{u}_{t+k}$ & $d$            & EMA target control context (Sec.~\ref{sec:method_control_jepa}) \\
\bottomrule
\end{tabular}
\label{tab:notation}
\end{table}

\paragraph{Token-count conservation and the Eq.~\ref{eq:token_update} update.}
A point that is implicit in Sec.~\ref{sec:method} and worth making explicit: every memory layer maps $N$
tokens to $N$ tokens, so the token count is conserved across all $L$ layers. The only operation
that mixes a per-token quantity with a single per-timestep vector is the inter-layer update
(Eq.~\ref{eq:token_update}),
\begin{equation}
Z^{\ell+1}_t \;=\; \bar{Z}^\ell_t \;+\; f^\ell_{\text{upd}}\!\big([E^\ell_t,\, h^\ell_t]\big),
\tag{11}
\end{equation}
where $\bar{Z}^\ell_t \in \mathbb{R}^{N\times d}$ and $E^\ell_t \in \mathbb{R}^{N\times d}$ are
per-token, but $h^\ell_t \in \mathbb{R}^{d}$ is a single working-state vector. We resolve this by
\emph{broadcasting} $h^\ell_t$ across the $N$ token positions before concatenation: each trace
token $e^\ell_{t,i}$ is paired with the same $h^\ell_t$, so for token position $i$,
\begin{equation}
\big(Z^{\ell+1}_t\big)_i \;=\; \big(\bar{Z}^\ell_t\big)_i \;+\;
f^\ell_{\text{upd}}\!\big([\,e^\ell_{t,i},\, h^\ell_t\,]\big),
\qquad i = 1,\dots,N,
\end{equation}
where $f^\ell_{\text{upd}}:\mathbb{R}^{2d}\!\to\!\mathbb{R}^{d}$ is a shared per-token MLP applied
independently at every position. Thus the recalled, consolidated working state $h^\ell_t$ is
written back into all event tokens, while the output remains $N\times d$ and feeds the next layer
unchanged in token count.

\subsection{Control-Indexed Memory Cell and Control-JEPA}
\label{app:cell}

We expand the four operations of each memory layer (Sec.~\ref{sec:method_memory}), then the prospective objective (Sec.~\ref{sec:method_control_jepa}) and the action head (Sec.~~\ref{sec:method_policy}).

\paragraph{Event binding (Eq.~\ref{eq:event_mixer}).}
The binding block $\mathrm{Mixer}^\ell$ is a residual Transformer-style token mixer applied
\emph{within a single timestep} over the $N$ event tokens, using 8 attention heads and
pre-LayerNorm. It first performs multi-head self-attention over the tokens of the current
frame and then applies a residual feed-forward sublayer; it does not attend across time.
Because binding precedes temporal propagation, each written event already contains task- and
body-conditioned evidence: visual, proprioceptive, and language tokens reinterpret one another
before any trace is formed. Binding is therefore causal by construction, as it operates only on
tokens of the current timestep.


\paragraph{Token-grounded trace propagation (Eq.~\ref{eq:slow_episodic}).}
Temporal memory is carried by a selective state-space layer $\mathrm{SSM}^\ell_{\text{slow}}$
(Mamba2-style selective SSM, latent width 512, state dimension 128, convolution width 4,
and expansion factor 1) applied \emph{token-wise}: a single shared kernel processes each token
stream $i$ independently along the causal time axis,
\begin{equation}
e^\ell_{1:T,i} \;=\; \mathrm{SSM}^\ell_{\text{slow}}\!\big(\bar{Z}^\ell_{1:T,i}\big),
\qquad E^\ell_t = \{e^\ell_{t,i}\}_{i=1}^{N}.
\tag{6}
\end{equation}
Sharing one kernel across positions keeps the trace bank token-aligned: trace $i$ at every
timestep summarizes the causal history of token stream $i$. This is what lets the bank preserve
$N$ separable histories rather than collapsing them into one recurrent vector, realizing
\emph{separability}. The recurrence is strictly causal: $e^\ell_{t,i}$ depends only on
$\bar{Z}^\ell_{1:t,i}$.

\paragraph{Control index and control context (Eqs.~\ref{eq:control_seed}--\ref{eq:control_context}).}
For each timestep, the policy needs the trace that answers the current control question, not all evidence equally. The control index is formed from the bound proprioception and language tokens,
\begin{equation}
c^\ell_t = f^\ell_c\!\big([\bar{p}^\ell_t,\, \bar{q}^\ell_t]\big) \in \mathbb{R}^d,
\tag{7}
\end{equation}
where $f^\ell_c$ is a two-layer MLP with GELU nonlinearity followed by LayerNorm. We derive
the index from the body and task channels rather than directly from scene tokens: the body and
instruction specify \emph{what the current decision is}, whereas the present scene may be aliased
at $t_d$ and can make the query ambiguous. The present scene re-enters in a controlled way through
a refinement step that attends from the index to the bound event tokens,
\begin{equation}
u^\ell_t = \mathrm{LN}\!\big(c^\ell_t + \mathrm{Attn}(c^\ell_t, \bar{Z}^\ell_t, \bar{Z}^\ell_t)\big),
\tag{8}
\end{equation}
using 8 attention heads and producing a control context $u^\ell_t$ that combines task, body, and
present-scene evidence before addressing the trace bank.


\paragraph{Control-indexed recall and working-state consolidation (Eqs.~\ref{eq:episodic_recall}--\ref{eq:working_state}).}
The control context recalls evidence by attending to the token-grounded traces, and the result is fused into a fast working state:
\begin{align}
r^\ell_t &= \mathrm{LN}\!\big(\mathrm{Attn}(u^\ell_t, E^\ell_t, E^\ell_t)\big), \tag{9}\\
w^\ell_t &= f^\ell_w\!\big([u^\ell_t, r^\ell_t]\big), \qquad
h^\ell_{1:T} = \mathrm{SSM}^\ell_{\text{fast}}(w^\ell_{1:T}). \tag{10}
\end{align}
Here $\mathrm{MHA}$ uses 8 attention heads, and $f^\ell_w$ is a two-layer MLP with GELU
nonlinearity and LayerNorm. Because the keys and values are the traces $E^\ell_t$ and the query is the current control context, the same bank yields different recalled content under different decision states---recall is \emph{addressable} rather than purely similarity-based. The fast stream $\mathrm{SSM}^\ell_{\text{fast}}$ is a Mamba2-style selective SSM with latent width 512, state dimension 32, convolution width 4, and expansion factor 2; it maintains the policy-facing working state, in contrast to the slow stream that maintains event evidence. The trace bank is then written back to the event tokens via Eq.~\ref{eq:token_update}, and stacking $L$ layers repeats binding, propagation, recall, and consolidation; the final policy-facing state is $h_t = h^L_t$.

\paragraph{Prospective training with Control-JEPA (Eqs.~\ref{eq:jepa_predictor}--\ref{eq:control_jepa}).}
Control-JEPA is our control-conditioned adaptation of the joint-embedding predictive principle
of JEPA~\cite{assran2023selfsupervised}; it is not the method applied directly, but a reuse of its predict-in-representation-space idea with the prediction target being \emph{future control context} rather
than image features. During training, an EMA target branch encodes future embodied event tokens
into target control contexts $\tilde{u}_{t+k}$ using the same Eqs.~\ref{eq:control_seed}--\ref{eq:control_context}) under stop-gradient. This branch is discarded at inference. For each horizon $k\in K$, a predictor reads the causal
working state and a horizon embedding,
\begin{equation}
\hat{u}_{t+k} = g_\theta\!\big([h_t, \eta_k]\big),
\tag{12}
\end{equation}
where $g_\theta$ is a two-layer MLP with GELU nonlinearity. The objective is
\begin{equation}
\mathcal{L}_{\text{CJEP}} =
\frac{1}{\sum_{k\in K'}\lambda_k}
\sum_{k\in K'} \lambda_k\, \rho\!\big(\hat{u}_{t+k}, \mathrm{sg}(\tilde{u}_{t+k})\big)
+ \lambda_{\text{var}} \mathcal{L}_{\text{var}},
\tag{13}
\end{equation}
where $K'$ denotes horizons valid for the current sequence, $\rho$ is a smooth-$L_1$ alignment
loss, and $\mathrm{sg}(\cdot)$ stops gradients through the EMA target. The variance term prevents
representation collapse; we use the per-dimension hinge
$\mathcal{L}_{\text{var}} = \frac{1}{d}\sum_{j=1}^{d}\max(0,\, \gamma -
\mathrm{std}(\hat{u}^{(j)}))$, with target std $\gamma=1.0$ and
$\lambda_{\text{var}}=0.05$. The horizon set is
$K=\{1,2,4,8,16,32\}$ with weights
$\lambda_k=(1,1,1,0.5,0.5,0.25)$, fixed across experiments. Horizons beyond the remaining
episode length are skipped. The EMA target momentum is $0.99$ and is updated once per optimizer
step without warmup. Because each target $\tilde{u}_{t+k}$ is the context a later policy step
actually uses, Control-JEPA makes the working state prospective while leaving causal inference
unchanged: no future tokens enter the inference path.


\paragraph{Memory-conditioned action head (Eqs.~\ref{eq:policy_context}--\ref{eq:flow_loss}).}
The final state conditions a transformer rectified-flow head through policy tokens
$C_t = W_h h_t$ (Eq.~\ref{eq:policy_context}). We train with a \emph{clean-endpoint} ($x_1$-prediction)
parameterization rather than the velocity-field form: given the normalized ground-truth horizon
$A_{t:t+H-1}$, we sample $A_0\sim\mathcal{N}(0,I)$, $\tau\sim\mathcal{U}(0,1)$, form
$A_\tau=(1-\tau)A_0+\tau A_{t:t+H-1}$ (Eq.~\ref{eq:flow_interpolant}), and regress the clean endpoint
\begin{equation}
\hat{A}_{t:t+H-1} = F_\theta(A_\tau, \tau, C_t),
\qquad
\mathcal{L}_{\text{act}} = \mathbb{E}\big[\,\|\hat{A}_{t:t+H-1} - A_{t:t+H-1}\|_2^2\,\big].
\tag{16}
\end{equation}
At inference, an action chunk is initialized from the configured source state (Gaussian noise, or zero for deterministic sample-mode inference) and updated along the rectified-flow path using the predicted endpoint over 50 Euler integration steps. The full objective is
$\mathcal{L} = \mathcal{L}_{\text{act}} + \alpha \mathcal{L}_{\text{CJEP}}$ (Eq.~17), with $\alpha=0.05$ in the main experiments (value in Table~\ref{tab:hparams}).


\section{Experimental Setup}
\label{app:setup}

\subsection{Simulation Benchmarks}
\label{app:sim}

We evaluate our method on three public benchmarks. For each benchmark, we strictly follow the protocol defined by the corresponding cited source, ensuring that our results are directly comparable with the reported baselines. Any deviations from the original protocol are explicitly stated shown in Table~\ref{tab:sim_protocol}.

\begin{table}[h]
\centering
\setlength{\tabcolsep}{4pt}
\caption{\textbf{Simulation benchmark protocols.} Configuration used for each public
benchmark. ``Policy'' indicates whether a single mixed-task policy or one policy per
task is trained. ``Protocol'' names the source whose evaluation setup we match.}
\resizebox{\textwidth}{!}{
\begin{tabular}{l c c c c c l}
\toprule
\textbf{Benchmark} & \textbf{\#Tasks} & \textbf{Policy} & \textbf{\#Demos/task} & \textbf{\#Rollouts/task} & \textbf{\#Seeds} & \textbf{Protocol followed} \\
\midrule
LIBERO-10    & 10 & mixed          & 50 & 50 & 3 & LIBERO~\cite{liu2023libero} \\
MemoryBench  & 3  & task-specific  & 100 & 25 & 4 & SAM2Act~\cite{fang2025sam2act} \\
MIKASA-Robo (mixed)    & 5 & mixed         & 250 & 30 & 3 & MIKASA-Robo~\cite{cherepanov2025memory} \\
MIKASA-Robo (special.) & 2 & task-specific & 250 & 30 & 3 & DP-VPWEM~\cite{lei2026vpwem} \\
\bottomrule
\end{tabular}
}
\label{tab:sim_protocol}
\end{table}

\paragraph{LIBERO-10.}
LIBERO-10 contains 10 long-horizon, language-conditioned tasks. We train a
single mixed-task policy on 50 demonstrations per task
(500 demonstrations in total) and evaluate with 50 rollouts per task. Results are averaged over 3 seeds; we report the mean $\pm$ standard deviation across seeds.

\paragraph{MemoryBench.}
MemoryBench evaluates spatial memory in manipulation. We train 3 task-specific policies (one per task), each on 100 demonstrations, and evaluate with 25 rollouts per task over 4 seeds.
%
We follow the standard MemoryBench protocol introduced by SAM2Act~\cite{fang2025sam2act}, under which the published baselines are RVT-2 (54.0\%), SAM2Act (55.0\%), and SAM2Act+ (94.3\%). Under this protocol, Chameleon reaches 97.3$\pm$4.5\%.
We note that ReMem-VLA reports 94.5\% under a \emph{modified} protocol. Since this protocol differs from ours, the 94.5\% result is only partially comparable. We therefore include it for context rather than as a like-for-like baseline, especially because ReMem-VLA uses a substantially larger model than ours.

%

\paragraph{MIKASA-Robo.}
MIKASA-Robo is a simulated benchmark of non-Markovian tasks. We report two settings, matching the two rows in Table~\ref{tab:public_results}: a 5-task mixed policy and a 2-task specialist setting. 
For the mixed setting, we train on 250 demonstrations per task and evaluate with 30 rollouts per task over 3 seeds. For the specialist setting, we use the two-task subset reported by VPWEM~\cite{lei2026vpwem}, train one policy per task with 250 demonstrations, and evaluate with 30 rollouts per task.

\paragraph{Baseline scores.}
All baseline numbers in Table~\ref{tab:public_results} are taken from the cited sources under the same
protocol as ours unless noted. When a score is quoted from a secondary comparison
table rather than the original paper, we cite both the original method and the
reporting source, following the convention in the Table~\ref{tab:public_results} footnote.

\subsection{Real-Robot Camo-Dataset}
\label{app:camo}

\begin{figure}[t]
\centering
\includegraphics[width=\textwidth,trim=0bp 200bp 200bp 0bp,clip]{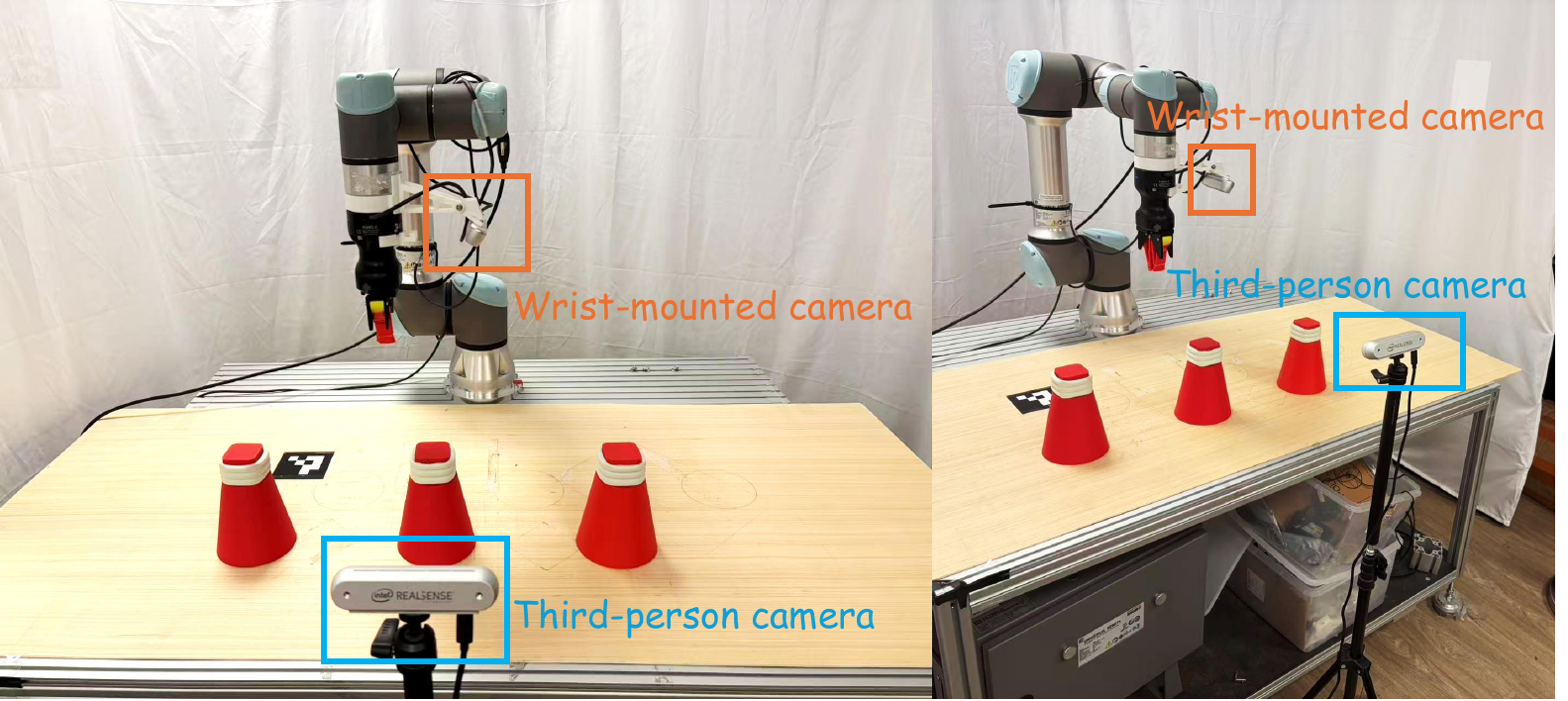}
\caption{\textbf{Real-robot setup for \benchname.} UR5 workspace with the
\(2\) camera viewpoints annotated.}
\label{fig:robot_setup}
\end{figure}

\paragraph{Hardware.}
All real-robot experiments use a 6-DoF UR5 arm with a Robotiq Hand-E adaptive parallel-jaw end-effector. We record 2 RGB camera views: one wrist-mounted and one third-person. Images are captured at $224^2$ resolution, and the policy runs at a control frequency of 30 (stride 4 $\rightarrow$ 7.5) Hz. 

\paragraph{Data collection.}
Demonstrations are collected via the leader--follower teleoperation method. We collect $120$ demonstrations per task. Episode lengths differ across tasks: \serveplate{} [614--1078], \shellgame{} [838--1224], and \addpeasoning{} [about 1100--2300] steps, reflecting their differing horizons (\addpeasoning{} is longest, as it repeats the seasoning subgoal three times). Because each task is a memory-dependent decision among a fixed set of equiprobable options, we balance demonstrations across the latent variable $z$ so that no option is over-represented: a policy must therefore recover $z$ from history rather than exploit a label prior. Concretely, for \serveplate{} and \shellgame{}, the latent has three equiprobable values ($p_e=1/3$; cf.\ Table~\ref{tab:memory_taxonomy}), so the target plate and the object-containing cup are each the correct answer in $1/3$ of demonstrations ([40] demonstrations per value at $120$ per task). For \addpeasoning{}, the latent is a length-three ordering over subgoals ($p_e=1/27$), and we sample the orderings uniformly across demonstrations. This balancing makes chance-level decision accuracy exactly $p_e$, so that any DSR above $p_e$ reflects history recovery rather than a dataset bias toward a frequent option.

\begin{table}[h]
\centering
\setlength{\tabcolsep}{4pt}
\caption{\textbf{Task design in \benchname.} The dataset contains three real-robot tasks that instantiate delayed control relevance through complementary hidden variables. $p_e$ is the chance decision rate used for chance-adjusted DSR. For \addpeasoning, DSR is scored at the complete sequence level.}
\resizebox{\textwidth}{!}{
\begin{tabular}{>{\raggedright\arraybackslash}p{2.2cm} >{\raggedright\arraybackslash}p{3.0cm} >{\raggedright\arraybackslash}p{3.5cm} >{\raggedright\arraybackslash}p{3.5cm} c}
\toprule
\textbf{Task} & \textbf{Hidden episode variable} & \textbf{Aliased decision point} & \textbf{Diagnostic role} & \textbf{$p_e$} \\
\midrule
\textbf{\serveplate} & Which visually similar plate was used or contaminated earlier. & Candidate plates appear interchangeable when the robot selects the target. & Event-object binding under delayed relevance. & $1/3$ \\
\textbf{\shellgame} & The hidden object's location after occlusion and swaps. & Cups are visually identical after rearrangement. & Spatial tracking under occlusion and distractor motion. & $1/3$ \\
\textbf{\addpeasoning} & Which subgoals have already been completed in a repeated sequence. & The workspace returns to a similar state after each seasoning action. & Sequential progress memory and prevention of repetition or omission. & $1/27$ \\
\bottomrule
\end{tabular}
}
\label{tab:memory_taxonomy}
\end{table}

\paragraph{Tasks and latent variable $z$.}
Each task contains an early evidence event whose relevance is delayed until a later,
visually aliased decision point. We summarize the latent variable and the source of
aliasing per task:
\begin{itemize}
    \item \textbf{Clean a specified plate.} $z$ is determined by which plate the human interacts with and places down as the target plate.
    Aliasing arises because, after the cue, the human interaction has ended and the decision frame contains three visually similar plates, any of which could have been the target.
  \item \textbf{Play shell game.} $z=$ the object-containing cup. Aliasing
  arises because the cups are shuffled and the ball is occluded, so all cups look identical at decision time.
  \item \textbf{Add various seasonings.} $z=$ the set of completed subgoals.
  Aliasing arises because the scene after each addition looks the same, so the next correct seasoning depends only on history.
\end{itemize}
Formally, at the annotated decision time $t_d$ the visible scene satisfies
$O_{t_d}(z_i)\approx O_{t_d}(z_j)$ while the history-correct action differs,
$a^\star(z_i)\neq a^\star(z_j)$ (Eq.~\ref{eq:aliasing_condition}).

\paragraph{Annotation of aliased decision times $t_d$.}
Two human annotators annotate the decision time $t_d$ for each episode as the first frame after the evidence event for $z$ has ended and the robot arm has reached the task-specific canonical decision pose. Concretely, this corresponds to the frame after the human has placed down the target plate and withdrawn in \serveplate{}, after the cup shuffle has finished and the ball is fully occluded in \shellgame{}, and after the previous seasoning addition has been completed in \addpeasoning{}. At this frame, the robot is about to choose among the candidate plates, cups, or seasonings, respectively, while neither the current image nor proprioception reveals the correct choice. These annotations are used only for evaluation and not during training.


\begin{table}[t]
\centering
\scriptsize
\setlength{\tabcolsep}{4pt}
\renewcommand{\arraystretch}{0.88}
\caption{\textbf{Parameter budget of Chameleon.} Breakdown by component. The text encoder is frozen during training. ``Trainable'' excludes it. The ``$\sim$60M'' claim in the main text refers to the trainable-parameter budget, not the total parameter count.}
\resizebox{0.7\textwidth}{!}{
\begin{tabular}{l c c}
\toprule
\textbf{Component} & \textbf{\#Params} & \textbf{Trainable?} \\
\midrule
Visual encoder (DP-style patch encoder) & 22.9M & yes \\
Text encoder (DistilBERT)               & 66.4M & frozen \\
Memory layers ($L$ control-indexed cells) & 20.3M & yes \\
\quad -- SSM$_{\text{slow}}$ / SSM$_{\text{fast}}$ & 5.1M & yes \\
\quad -- Mixer / control index / attention        & 15.2M & yes \\
Rectified-flow action head              & 23.1M & yes \\
Projections (proprio, language, null token) & 0.4M & yes \\
\midrule
\textbf{Trainable total}                & $\sim$66.7M & --- \\
\textbf{Frozen total}                   & $\sim$66.4M & --- \\
\textbf{Overall total}                  & $\sim$133.1M & --- \\
\bottomrule
\end{tabular}
}
\label{tab:param_budget}
\end{table}

\begin{table}[t]
\centering
\scriptsize
\setlength{\tabcolsep}{3pt}
\renewcommand{\arraystretch}{0.92}
\caption{
\textbf{Reported model scale.}
We report approximate total/backbone model scale, not benchmark-specific trainable parameters.
Ranges indicate commonly reported configurations when exact counts are implementation-dependent.
}
\resizebox{0.7\textwidth}{!}{
\begin{tabular}{l c c}
\toprule
\textbf{Model} & \textbf{Approx. scale} & \textbf{Model type} \\
\midrule
\textbf{\modelname{} (ours)} & \textbf{100–500M} & policy model \\
\midrule
\DP~\cite{chi2023diffusionpolicy} & 100--500M & policy model \\
ACT~\cite{zhaolearning} & 10--100M & policy model \\
Flow Matching~\cite{lipman2022flow,liu2023flowstraight} & 100--500M & policy model \\
DP-T / DP-CNN~\cite{chi2023diffusionpolicy,reuss2025efficient} & 100--500M & policy model \\
QueST~\cite{reuss2025efficient} & 100--500M & policy model \\
MoDE~\cite{reuss2025efficient} & 100--500M & policy model \\
MaIL~\cite{jia2024mail} & 10--100M & policy model \\
DP-VPWEM~\cite{lei2026vpwem} & 100--500M & policy model \\
GMP~\cite{gao2026gatedmemorypolicy} & 100--500M & policy model \\
\midrule
RVT-2~\cite{goyal2024rvt2} & 100--500M & 3D policy model \\
SAM2Act / SAM2Act+~\cite{fang2025sam2act} & 100--500M & 3D policy model \\
\midrule
CronusVLA-small~\cite{li2025cronusvla} & 500M & VLA model \\
$\pi_0$~\cite{black2024pi_0} & 3.3B & VLA model \\
OpenVLA~\cite{kim2024openvla} & 7B & VLA model \\
OpenVLA-OFT~\cite{kim2025openvlaoft} & 7B & VLA model \\
CoT-VLA~\cite{zhao2025cot} & 7B & VLA model \\
CronusVLA~\cite{li2025cronusvla} & 7B & VLA model \\
TriVLA~\cite{liu2025trivla} & 7B & VLA model \\
4D-VLA~\cite{zhang20264d} & 7B & VLA model \\
SpatialVLA~\cite{qu2025spatialvla} & 7B & VLA model \\
MemoryVLA~\cite{shi2026memoryvla} & 7.3B & VLA model \\
ReMem-VLA~\cite{li2026rememvla} & 7B & VLA model \\
\bottomrule
\end{tabular}
}
\label{tab:param_compare}
\vspace{-0.6em}
\end{table}
\subsection{Evaluation Metrics}
\label{app:metrics}

We score execution and memory-dependent choice separately (main text Eq.~\ref{eq:metrics_basic}.
Let $M$ denote a scorable manipulation and $D$ denote a history-correct choice:
\begin{equation*}
\mathrm{MSR}=P(M),\qquad \mathrm{DSR}=P(D\mid M),\qquad \mathrm{SR}=P(M\cap D).
\end{equation*}

\paragraph{Scoring rubric.}
\begin{itemize}
    \item \textbf{$MSR$.} A rollout counts as a scorable manipulation if
    the robot completes the task-specific manipulation procedure on a physically valid candidate target, without execution failure such as collision, dropping the object, or failing to complete the required pick-and-place sequence, regardless of whether the chosen target is history-correct.
    \item \textbf{$DSR$.} Given $M=1$, the decision is correct if the selected candidate matches the history-correct target or subgoal determined by $z$.
\end{itemize}
Scoring is performed independently by two raters.

\paragraph{Number of trials.}
For each method, we evaluate 36 rollouts per task and report the average over the three tasks.


\section{Implementation Details of Chameleon}
\label{app:impl}

\subsection{Parameter Budget}
\label{app:params}

Table~\ref{tab:param_budget} breaks down the parameter count of Chameleon by component.
The total includes the frozen DistilBERT text encoder. 

\paragraph{Comparison to baselines.}
Table~\ref{tab:param_compare} situates Chameleon against the baselines we compare to,
supporting the claim that Chameleon is state-of-the-art \emph{among same-size models}
while outperforming substantially larger VLA systems.


\subsection{Hyperparameters}
\label{app:hparams}

Table~\ref{tab:hparams} lists all hyperparameters. Architecture and optimization settings
are \emph{shared across all benchmarks}; only the input/output configuration (camera count,
resolution, action dimension, horizon) varies per benchmark, as the benchmarks use different
robots and sensor layouts. Notably, the loss weight $\alpha$ (Eq.~\ref{eq:total_loss}) and the Control-JEPA horizon set $K$ are \emph{fixed across all experiments} and are not tuned per benchmark.

\begin{table}[h]
\centering
\setlength{\tabcolsep}{6pt}
\caption{\textbf{Hyperparameters of Chameleon.} Top block: model settings shared across
benchmarks; optimization rows that vary are reported in the order
LIBERO-10 / MemoryBench / MIKASA / Camo. Bottom block: input/output configuration.
$\alpha$ and $K$ are fixed across all experiments.}
\begin{tabular}{l l}
\toprule
\multicolumn{2}{l}{\textbf{Shared model and training hyperparameters}} \\
\midrule
Feature dimension $d$            & 512 \\
Memory layers $L$                & 2 \\
Mixer attention heads            & 8 \\
SSM$_{\text{slow}}$ state dim    & 128 \\
SSM$_{\text{fast}}$ state dim    & 32 \\
RF head depth / heads            & 6 / 8 \\
RF inference steps               & 50 \\
\midrule
Optimizer                        & AdamW \\
Learning rate                    & $1\times10^{-4}$ \\
LR schedule / warmup             & cosine / 100 steps \\
Weight decay                     & $1\times10^{-6}$ \\
Batch size                       & 64 effective \\
Training steps                   & 100k optimizer steps \\
Gradient clip                    & 10 \\
Precision                        & bf16 \\
Hardware                         & $1\times$ NVIDIA GeForce RTX 5090 \\
\midrule
Action loss weight (implicit)    & $1$ \\
Control-JEPA weight $\alpha$ (Eq.~17)   & 0.05 \\
Variance weight $\lambda_{\text{var}}$ (Eq.~13) & 0.05 \\
Horizon weights $\{\lambda_k\}$  & $\{1.0, 1.0, 1.0, 0.5, 0.5, 0.25\}$ \\
Control-JEPA horizon set $K$     & $\{1, 2, 4, 8, 16, 32\}$ \\
EMA target momentum              & 0.99 \\
\midrule\midrule
\multicolumn{2}{l}{\textbf{Per-benchmark input/output configuration}} \\
\midrule
& LIBERO-10 / MemoryBench / MIKASA / Camo \\
Camera views $V$        & 2 / 2 / 2 / 2 \\
Image resolution        & $224^2$ / $224^2$ / $128^2$ / $224^2$ \\
Visual tokens per view $N_v$ & 36 / 36 / 16 / 36 \\
Proprioception dim      & 8 / 10 / 25 / 10 \\
Action dimension        & 7 / 8 / 8 / 10 \\
Action horizon $H$      & 16 / 16 / 8 / 16 \\
Control frequency (Hz)  & 20 / -- / 10 / 30 (stride 4 $\rightarrow$ 7.5) \\
\bottomrule
\end{tabular}
\label{tab:hparams}
\end{table}

\paragraph{Frozen and learned components.}
The DistilBERT text encoder is frozen throughout training. The image encoders, visual token
projection layers, proprioception projection, language projection, memory layers, Control-JEPA
predictor, and action head are learned end to end. The EMA target branch used by Control-JEPA is
a stop-gradient copy of the online control-context path and is updated only by exponential moving
average. When language is enabled but an instruction is missing, we encode an empty instruction
string with the frozen text encoder and map it through the learned language projection; when
language is disabled, no separate language token is inserted.


\section{Per-Task Results on Public Benchmarks}
\label{app:pertask}

Table~\ref{tab:pertask} reports per-task success rates underlying the averages in Table~\ref{tab:public_results}.
For each benchmark we follow the baseline-selection convention: where Chameleon is
state-of-the-art we list the two strongest published baselines; where it is not, we list
the strongest baseline overall together with the strongest baseline that Chameleon
surpasses, so that Chameleon's standing is shown rather than obscured by the average.

\begin{table}[h]
\centering
\setlength{\tabcolsep}{5pt}
\renewcommand{\arraystretch}{1.15}
\caption{\textbf{Per-task results on public benchmarks.} Success rate (\%). Each block uses the baselines selected per the convention in the text. Bold = best in block.}
\resizebox{\textwidth}{!}{
\begin{tabular}{l c c c}
\toprule
\textbf{Task} & \textbf{Chameleon (ours)} & \textbf{DP-family baseline} & \textbf{VLA baseline} \\
\midrule
\multicolumn{4}{l}{\textit{\textbf{MemoryBench~\cite{fang2025sam2act}}} \;--\; SOTA: 2 strongest baselines (SAM2Act+~\cite{fang2025sam2act}, ReMem-VLA~\cite{li2026rememvla})} \\
\midrule
Reopen Drawer  & 92.0$\pm$13.5 & 84.0$\pm$0.0 & 100.0 \\
Put Block Back  & 100.0$\pm$0.0 & 100.0$\pm$0.0 & 93.0 \\
Rearrange Block  & 100.0$\pm$0.0 & 99.0$\pm$2.0 & 99.0 \\
\textit{Average}      & 97.3$\pm$4.5 & 94.3 & 94.5 \\
\midrule
\multicolumn{4}{l}{\textit{\textbf{MIKASA-Robo~\cite{cherepanov2025memory}}} (mixed, 5-task) \;--\; SOTA: 2 strongest baselines (GMP~\cite{gao2026gatedmemorypolicy}, MemoryVLA~\cite{shi2026memoryvla})} \\
\midrule
Intercept Medium  & 58.9$\pm$2.5 & -- & 24.0 \\
Remember Color 3  & 94.4$\pm$1.3 & -- & 44.0 \\
Remember Color 5  & 72.2$\pm$1.7 & -- & 30.0 \\
Remember Color 9  & 53.3$\pm$1.6 & -- & 20.0 \\
Shell Game  & 96.7$\pm$1.4 & -- & 88.0 \\
\textit{Average}      & 75.1$\pm$1.4 & 67.8 & 41.2 \\
\bottomrule
\end{tabular}
}
\label{tab:pertask}
\end{table}

\paragraph{LIBERO-10.}
LIBERO-10 does not report per-task scores for all compared baselines, so we analyze the task-level behavior of \modelname{} rather than making per-task baseline comparisons. The most common failures of \modelname{} are low-level manipulation errors, such as unstable grasping or imprecise placement, rather than failures to retain task-relevant history. This suggests that the remaining gap on LIBERO-10 is mainly due to execution accuracy and manipulation robustness, while the proposed memory mechanism still preserves competitive average performance with a much smaller parameter budget.

\paragraph{MemoryBench.}
On MemoryBench, \modelname{} achieves the best average performance among the compared methods, but its advantage is not uniform across tasks. It matches or exceeds the strongest baselines on Put Block Back and Rearrange Block, where performance is nearly saturated. The lower score on Reopen Drawer is mainly caused by execution sensitivity: the drawer button is very low in the benchmark setup, and the policy often produces the intended pressing motion but fails to press it fully. Thus, the remaining error is more closely tied to fine contact execution than to memory retrieval.

\paragraph{MIKASA-Robo (mixed).}
On MIKASA-Robo, the strongest evidence for non-Markovian memory comes from the Remember Color tasks, where the correct action depends on retaining earlier color information that is no longer directly available at decision time. \modelname{} improves substantially over MemoryVLA on all three Remember Color variants, with gains of 50.4, 42.2, and 33.3 percentage points for Remember Color 3, 5, and 9, respectively. The gain on Shell Game is smaller because this benchmark instance is relatively simple and the VLA baseline is already strong. Intercept Medium is less memory-demanding and is closer to a reaction-time task, so we treat its improvement as supporting evidence for general policy quality rather than as the main evidence for non-Markovian memory.


\section{Full Ablations with Statistics}
\label{app:ablation}

Table~\ref{tab:ablation_full} re-reports the ablation study of main-text Table~1 with
confidence intervals and an explicit mapping from each removed mechanism to the functional
property it targets. Each ablation isolates one of the three design principles: token-grounded
traces realize \emph{separability}, the control index realizes \emph{addressability}, and
Control-JEPA realizes \emph{prospectiveness}.

\begin{table}[h]
\centering
\setlength{\tabcolsep}{4pt}
\renewcommand{\arraystretch}{1.15}
\caption{\textbf{Ablation study on Camo-Dataset.} Avg.\ columns aggregate over tasks. The ``Ablated property'' column names the functional capability each variant removes.}
\resizebox{\textwidth}{!}{
\begin{tabular}{l c c c c c}
\toprule
\textbf{Method} & \textbf{Clean plate} & \textbf{Shell game} & \textbf{Seasoning} & \textbf{Avg.\ DSR / SR} & \textbf{Ablated property} \\
 & DSR/SR & DSR/SR & DSR/SR & & \\
\midrule
\textbf{Chameleon (full)} & 91.2/86.1 & 86.1/86.1 & 65.2/41.7 & 80.8 / 71.3 & --- \\
\midrule
w/o memory               & 26.7/22.2 & 34.4/30.6 & 0.0/0.0  & 20.4 / 17.6 & separability (all memory) \\
similarity retrieval bank & 41.4/33.3 & 28.6/22.2 & 0.0/0.0  & 23.3 / 18.5 & addressability (vs.\ control-indexed) \\
Vanilla Mamba memory      & 27.6/22.2 & 30.0/25.0 & 50.0/19.4 & 35.9 / 22.2 & separability (token-grounded traces) \\
w/o control index         & 40.7/30.6 & 45.8/30.6 & 60.0/16.7 & 48.8 / 26.0 & addressability \\
w/o Control-JEPA          & 82.8/66.7 & 71.0/61.1 & 61.1/30.6 & 71.6 / 52.8 & prospectiveness \\
\bottomrule
\end{tabular}
}
\label{tab:ablation_full}
\end{table}

\paragraph{Separability.}
Separability is the ability to keep visually similar histories distinguishable after the current observation becomes aliased. In Chameleon, this is implemented by writing embodied event tokens and propagating token-grounded slow traces, rather than compressing history into a single recurrent state. Removing memory entirely and replacing token-grounded traces with a vanilla Mamba state both reduce performance, indicating that delayed evidence is more useful
when preserved as separable token-grounded traces than when ignored or compressed into a
single recurrent state.

\paragraph{Addressability.}
Addressability is the ability to retrieve the trace required by the current control decision, rather than the most recent or visually most similar trace. In Chameleon, this is implemented by the control index and control-context query over the token-grounded trace bank. Both the similarity-retrieval bank and the removal of the control index fall well below the full model, indicating that recall is most useful when conditioned on the current control state rather than on visual similarity or undirected history access.

\paragraph{Prospectiveness.}
Prospectiveness is the ability to convert recalled evidence into a working state that supports upcoming action, rather than merely encoding what happened before. In Chameleon, this is encouraged by Control-JEPA, which trains $h_t$ to predict future control context. Removing Control-JEPA retains part of the decision signal, but reduces
SR from 71.3 to 52.8, indicating that predicting future control context
shapes $h_t$ into a working state that is more reliable for downstream action.

\paragraph{On the all-zero seasoning baselines.}
The matched imitation baselines (Diffusion Policy, ACT, Flow Matching) all score 0.0/0.0 on Add various seasonings. We emphasize that this reflects a
\emph{memory-mechanism} limitation rather than an unsolvable task: the Vanilla Mamba variant,
which adds only an undifferentiated recurrent memory, already reaches 50.0/19.4 DSR/SR
on the same task, and Chameleon reaches 65.2/41.7. \addpeasoning{} is scored at the
\emph{complete-sequence} level (Table~\ref{tab:memory_taxonomy}), so a memoryless policy that cannot track which subgoals are already done fails the whole sequence. The non-zero scores of memory-equipped variants confirm the task is solvable and that the gap is attributable to
memory, not to task design.

\section{Mechanistic Probe Details}
\label{app:probes}


This section describes how we compute the three probes in Sec.~\ref{sec:analysis} and Fig.~\ref{fig:analysis-completion}.
Each probe is evaluated at annotated aliased decision states in \benchname{}, where the current observation alone does not determine the correct action.
If an episode contains multiple such decision states, each state is counted as a separate probe sample.
Probe classifiers are trained and evaluated on disjoint episodes unless stated otherwise.

\subsection{Separability Probe (Fig.~~\ref{fig:analysis-completion}a)}

We decode the hidden episode variable $z$ from two representations: the current event tokens
$Z^0_t$ (``cur'') and the final policy-facing memory state $h^L_t$ (``mem'').
The decoder is a linear probe trained on annotated aliased decision states from the training split
and evaluated on held-out episodes with no episode overlap.
The held-out probe set contains 30 aliased decision states for \shellgame{} and 270 for
\addpeasoning{}.
Chance accuracy equals the per-task chance rate $p_e$ from Table~\ref{tab:memory_taxonomy}
($1/3$ for \shellgame{} and $1/27$ for \addpeasoning{}), so reported gaps are read against chance.

\begin{table}[t]
\centering
\setlength{\tabcolsep}{8pt}
\caption{\textbf{Separability probe accuracy (\%).} Decoding the hidden variable $z$ from
current tokens vs.\ memory state at $t_d$. Chance is the per-task $p_e$.}
\begin{tabular}{l c c c}
\toprule
\textbf{Task} & \textbf{Chance} & \textbf{Current $Z^0_t$} & \textbf{Memory $h^L_t$} \\
\midrule
Play shell game         & 33.3 & 46.7 & 83.3 \\
Add various seasonings  & 3.7  & 37.4 & 98.5 \\
\bottomrule
\end{tabular}
\label{tab:probe_sep}
\end{table}

\noindent The qualitative panel for \serveplate{} (Fig.~\ref{fig:analysis-completion}a) shows 3D UMAP projections of memory trajectories, which remain organized by earlier plate identity after visual aliasing.

\subsection{Addressability Probe (Fig.~~\ref{fig:analysis-completion}b)}
We test whether the control query $u^\ell_t$ addresses the relevant trace in the trace bank $E^\ell_t$.
At the aliased decision time in \addpeasoning{}, we fix $u^\ell_t$ at the value produced by the original current observation and intervene on $E^\ell_t$ at the final layer.
The four interventions are:
\begin{itemize}
  \item \textbf{Full}: original history, no edit.
  \item \textbf{Swap}: the relevant trace is replaced with a trace from another episode with a different $z$.
  \item \textbf{Mask+}: the relevant trace is removed.
  \item \textbf{Mask-}: irrelevant traces are removed, while the relevant trace is kept.
\end{itemize}
We define counterfactual choice accuracy as whether the selected subgoal matches the target implied by the active relevant trace after the intervention.
Under \textbf{Swap}, the active relevant trace is the swapped-in trace; thus, 87\% means that the policy follows the edited memory content rather than the original history.

\begin{table}[t]
\centering
\setlength{\tabcolsep}{10pt}
\caption{\textbf{Addressability counterfactual choice accuracy (\%).}
The control query $u^\ell_t$ is fixed while the trace bank $E^\ell_t$ is edited in \addpeasoning{}.}
\begin{tabular}{l c c c c}
\toprule
 & \textbf{Full} & \textbf{Swap} & \textbf{Mask+} & \textbf{Mask-} \\
\midrule
Choice accuracy & 93 & 87 & 40 & 90 \\
\bottomrule
\end{tabular}
\label{tab:probe_addr}
\end{table}

\noindent The pattern supports addressable recall: the choice remains accurate with the original trace, follows the swapped-in trace, degrades when the relevant trace is removed, and remains stable when irrelevant traces are removed.
This indicates that recall is driven by the trace selected by the control query, rather than by temporal proximity or visual similarity.


\subsection{Prospectiveness Probe (Fig.~\ref{fig:analysis-completion}c)}
We test whether the working state $h_t$ contains future action information before the action is executed.
At timesteps around each annotated aliased decision point, we train a lightweight decoder to predict a future control label from $h_t$.
For spatial-choice tasks, this label is the target endpoint selected later in the trajectory, such as the cup or plate endpoint that the end effector will act on.
For sequential tasks, it is the next annotated subgoal mode.
Fig.~\ref{fig:analysis-completion}(c) reports decoding accuracy over relative time, with the decision point at $0$.
Earlier high accuracy means that $h_t$ exposes future control information before it is needed.
The $+17$ and $+10$ annotations indicate how many timesteps earlier \modelname{} reaches the accuracy threshold than the w/o-Control-JEPA ablation.
The Trace Window baseline decodes from the trace bank without the prospective working state. These future control labels are used only for probing and training targets, not as privileged inputs at inference.

\clearpage

\end{document}